\documentclass{article}
\usepackage[nonatbib,preprint]{nips_2018}
\usepackage[ansinew]{inputenc}
\usepackage[T1]{fontenc}
\usepackage{hyperref}
\usepackage{color}
\usepackage{amsfonts}
\usepackage{nicefrac}
\usepackage{graphicx}
\usepackage{amsmath}
\usepackage{cite}

\input{tcilatex}
\begin{document}

\title{A Review of Tracking, Prediction and Decision Making Methods for
Autonomous Driving}
\author{ Florin Leon, Marius Gavrilescu \\
``Gheorghe Asachi'' Technical University of Ia\c{s}i, Romania\\
Faculty of Automatic Control and Computer Engineering \\
florin.leon@tuiasi.ro, mariusgav@tuiasi.ro 
}
\maketitle

\begin{abstract}
This literature review focuses on three important aspects of an autonomous
car system: tracking (assessing the identity of the actors such as cars,
pedestrians or obstacles in a sequence of observations), prediction
(predicting the future motion of surrounding vehicles in order to navigate
through various traffic scenarios) and decision making (analyzing the
available actions of the ego car and their consequences to the entire
driving context). For tracking and prediction, approaches based on (deep)
neural networks and other, especially stochastic techniques, are reported.
For decision making, deep reinforcement learning algorithms are presented,
together with methods used to explore different alternative actions, such as
Monte Carlo Tree Search.
\end{abstract}

\section{Introduction}

Autonomous car technology is already being developed by many companies on
different types of vehicles. Complete driverless systems are still at an
advanced testing phase, but partially automated systems have been around in
the automotive industry for the last few years. Autonomous cars have been
studied and developed by many universities, research centers and car
manufacturing companies, since the middle 1980s.

A famous competition was the DARPA Urban Challenge in 2007. Since then, many
self-driving-car competitions and trials have been performed. Relevant
examples include: the European Land-Robot Trial, which has been held from
2006 until the present, the Intelligent Vehicle Future Challenge, from 2009
to 2013, and the Autonomous Vehicle Competition, from 2009 to 2017. At the
same time, research on driverless cars has accelerated in both academia and
industry around the world.

In this review, we focus on three aspects of an autonomous car system:

\begin{itemize}
\item \textbf{Tracking:} assessing the identity of the actors (e.g., cars,
pedestrians, obstacles) in a sequence of observations. It is assumed that
some preprocessing of sensor data and/or input images has already been done;

\item \textbf{Prediction:} predicting the future motion of surrounding
vehicles in order to navigate through various traffic scenarios. Beside the
prediction of the simple physical behavior of the actors based on a set of
past observations, an important issue is to take into account their possible
interactions;

\item \textbf{Decision making}: analyzing the possible actions of the ego
car and their consequences for the entire driving context. It can be used
for the final trajectory planning of the vehicle.
\end{itemize}

\section{Tracking Methods}

Object tracking is an important part of ensuring accurate and efficient
autonomous driving. The identification of objects such as pedestrians, cars
and various obstacles from images and vehicle sensor data is a significant
and complex interdisciplinary domain, involving contributions from computer
vision, signal processing, machine learning, etc. Object tracking is an
essential part of ensuring safe autonomous driving, since it can aid in
obstacle avoidance, motion estimation, the prediction of the intentions of
pedestrians and other vehicles, as well as path planning. Most sensor data
that have to be processed take the form of point clouds, images, or a
combination of the two. Point cloud data may be handled in a multitude of
ways, the most common of which is some form of 3D grid, where a voxel engine
is used to traverse the point space. Some situations call for a
reconstruction of the environment from the point cloud which involves
various means of resampling and filtering. In some instances, stereo visual
information is available and disparities must be computed from the
left-right images. Stereo matching is not a trivial task and has the
drawback that many computations are required in order to ensure accuracy,
which usually has a significant impact on performance. In other cases,
multiple types of sensor data are available, thereby requiring registration,
point matching, image/point cloud fusion and many other such tasks. The
problem is further complicated by the necessity to account for temporal cues
and to estimate motion from time-based frames.

The scenes involved in autonomous driving scenarios rarely feature a single
individual target. Most commonly, multiple objects must be identified and
tracked concurrently, some of which may be in motion relative to the vehicle
and to each other. As such, most approaches in the related literature handle
more than one object and are therefore aimed at solving multiple object
tracking problems, commonly abbreviated as MOT.

The tracking problem can be summarized as follows: given a sequence of
sensor data from one or multiple vehicle-mounted acquisitions devices,
considering that several observations are identified in all or some of the
frames from the sequence, how can the observations from each frame be
associated with a set of objects (pedestrians, vehicles, various obstacles
etc.) and how can the trajectories of each such object be reconstructed and
predicted as accurately as possible?

Most related methods involve assigning an ID or identifying a response
for all objects detected within a frame and then attempting to match the IDs
across subsequent frames. This is often a complex task, considering that the
tracked objects may enter and leave the frame at different timestamps, they
may be occluded by the environment or may occlude each other, in addition to
the problems caused by defects in the acquired images such as noise,
sampling or compression artifacts, aliasing, acquisition errors etc.

Object tracking for automated driving most commonly has to operate on
real-time video. As such, the objective is to correlate tracked objects
across multiple video frames, in addition to individual object
identification. Accounting for variations in motion comes with an additional
set of pitfalls, such as when objects are affected by rotation or scaling
transformations, or when the movement speed of the objects is high relative
to the frame rate.

In the majority of cases, images are the primary modality for perceiving the
scene, as such a lot of efforts from the related literature are in the
direction of 2D MOT. The related algorithms are based on a succession of
detection and tracking, where consecutive detections which are similarly
classified are linked together to determine trajectories. A significant
challenge comes from the inevitable presence of noise in the acquired
images, which may adversely change the features of similar objects across
multiple frames. Consequently, the computation of robust features is an
important aspect of object detection. Features are representative of a wide
array of object properties: color, frequency and distribution, shape,
geometry, contours, various correlations that exist among the intensities of
segmented objects etc. Nowadays, the most popular means of feature detection
is using a supervised learning approach, where features start out as groups
of random values and are progressively refined using a machine learning
algorithm. Such an approach requires appropriate training data and a careful
selection of hyperparameters, often through trial-and-error. However, many
results from the related literature show that supervised classification and
regression methods offer the best results both in terms of accuracy and
robustness to affine transformations, occlusion and noise.

\subsection{Methods Using Neural Networks}

In terms of classifying objects from images, neural networks have seen a
steady rise in popularity in recent years, particularly the more elaborate
and complex convolutional and recurrent networks from the field of deep
learning. Neural networks have the advantage of being able to learn
important and robust features given training data that is relevant and in
sufficient quantity. Considering that a significant percentage of automotive
sensor data consists of images, convolutional neural networks (CNN) are
seeing widespread use in the related literature, for both classification and
tracking problems. The advantage of CNNs over more conventional classifiers
lies in the convolutional layers, where various filters and feature maps are
obtained during training. CNNs are capable of learning object features by
means of multiple complex operations and optimizations, and the appropriate
choice of network parameters and architecture can ensure that these features
contain the most useful correlations that are needed for the robust
identification of the targeted objects. While this choice is most often an
empirical process, a wide assortment of network configurations exist in the
related literature that are aimed at solving classification and tracking
problems, with high accuracies claimed by the authors. Where object
identification is concerned, in some cases the output of the fully-connected
component of the CNN is used, while in other situations the values of the
hidden convolutional layers are exploited in conjunction with other
filtering and refining methods.

\subsubsection{Learning Features from Convolutional Layers}

\begin{figure}[tbp]
\includegraphics[width={\textwidth}]{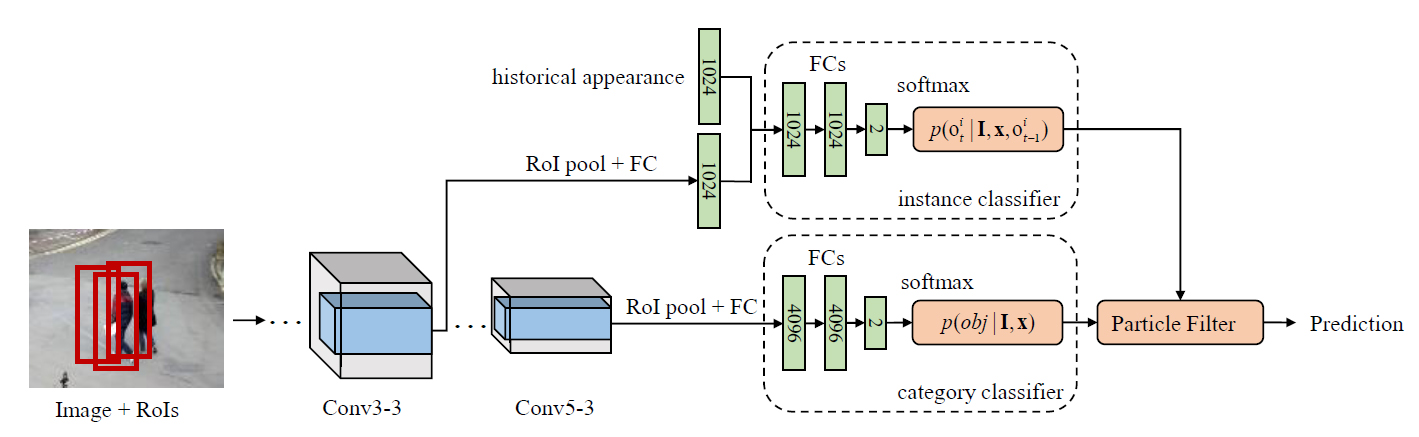}
\caption{An example of a CNN arrangement for MOT \protect\cite{8296360}}
\label{fig_8296360}
\end{figure}

Many results from the related literature systematically demonstrate that
convolutional features are more useful for tracking than other
explicitly-computed ones (Haar, FHOG, color labeling etc.). An example in
this sense is \cite{8296360}, which handles MOT using combinations of values
from convolutional layers located at multiple levels. The method is based on
the notion that lower-level layers account for a larger portion of the input
image and therefore contain more details from the identified objects, making
them useful, for instance, for handling occlusion. Conversely, top-level
layers are more representative of semantics and are useful in distinguishing
objects from the background. The proposed CNN architecture uses dual
fully-connected components, for higher and lower-level features, which
handle instance-level and category-level classification, respectively
(Figure \ref{fig_8296360}). The proper identification of objects,
particularly where occlusion events occur, involves the generation of
appearance models of the tracked objects, which can result from the
appropriate processing of the features learned within in CNN.

On a similar note, \cite{16Liu} notes that the output of the fully-connected
component of a CNN is not suitable for handling infrared images. Their
attempt to directly transfer CNNs pretrained with traditional images for use
with infrared sensor data is unsuccessful, since only the information from
the convolutional layers seems to be useful for this purpose. Furthermore,
the layer data itself requires some level of adaptation to the specifics of
infrared images. Typically, infrared data offers much less spatial
information than visual images, and is much more suited, for example, in
depth sensors for gathering distances to objects, albeit at a significantly
lower resolution compared to regular image acquisition. As such,
convolutional layers from infrared images are used in conjunction with
correlation filters to generate a set of weak trackers which provides
response maps with regard to the targets' locations. The weak trackers are
then combined in ensembles which form stronger response maps with a much
greater tracking accuracy. The response map of an image is, in general
terms, in an intensity image where higher intensities indicate a change or a
desired feature/shape/structure in the initial image, when exposed to an
operator or correlation filter of some kind. By matching or fusing responses
from multiple images within a video sequence, one could identify similar
objects (i.e. the same pedestrian) across the sequence and subsequently
construct their trajectories.

The potential of correlation filters is also exploitable for regular images.
These have the potential to boost the information extracted from the
activations of CNN layers, for instance in \cite{ConvDCF_ICCV15_VOTworkshop}%
, where the authors find that by applying the appropriate filters to
information drawn from shallow CNN layers, a level of robustness similar to
using deeper layers or a combination of multiple layers can be achieved. In 
\cite{ICIPReza2018_v3}, the authors also note the added robustness
obtainable by post-filtering convolutional layers. By using particle and
correlation filters, basic geometric and spatial features can be deduced for
the tracked objects, which, together with a means of adaptively generating
variable models, can be made to handle both simple and complex scenes.

An alternative approach can be found in \cite{iccv17b}, where discriminative
correlation filters are used to generate an appearance model from a small
number of samples. The overall approach is similar, involving feature
extraction, post-processing, the generation of response maps for carrying
out better model updates within the neural network. Contrary to other
similar results, the correlation filters used throughout the system are
learned within a one-layer CNN, which eventually can be used to make
predictions based on the response maps. Furthermore, residual learning is
employed in order to avoid model degradation, instead of the much more
frequently-used method of stacking multiple layers. Other tracking methods
learn a similar kind of mapping from samples in the vicinity of the target
object using deep regression \cite{deeplk-icra}, \cite{sensors-19-00387}, or
by estimating and learning depth information \cite{paper-deeptam}.

The authors of \cite{1608.03773CCOT} note that correlation filters have
limitations imposed by the feature map resolution and propose a novel
solution where features are learned in a continuous domain, using an
appropriate interpolation model. This allows for the more effective
resolution-independent compositing of multiple feature maps, resulting in
superior classification results.

Methods based on discriminative correlation filters are notoriously prone to
excessive complexity and overfitting, and various means are available for
optimizing the more traditional methods. The most noteworthy in this sense
is \cite{ECO1611.09224}, who employs efficient convolution operators, a
training sample distribution scheme and an optimal update strategy in an
attempt to boost performance and reduce the number of parameters. A
promising result which demonstrates significant robustness and accuracy is 
\cite{1510.07945MDnet}, who use a CNN where the first set of layers are
shared, as in a standard CNN; however at some point the layers branch into
multiple domain-specific ones. This approach has the benefit of splitting
the tracking problem into subproblems which are solved separately in their
respective layer sets. Each domain has its own training sequences and be
customized to can address a specific issue (such as distinguishing a target
with specific shape parameters from the background). A similar concept, i.e.
a network with components distinctly trained for a specific problem, can be
found in \cite{SANet}. In this case, multiple recurrent layers are used to
model different structural properties of the tracked objects, which are
incorporated into a parent CNN with the same purpose of improving accuracy
and robustness. The RNN layers generate what the authors refer to as
``structurally-aware feature maps'' which, when combined with pooled
versions of their non-structurally aware counterparts, significantly improve
the classification results.

\subsubsection{High-Level Features, Occlusion Handling and Feature Fusion}

Appearance models offer high-level features which are also used to account
for occlusion in much simpler and efficient systems, such as in \cite%
{1703.07402}, where computed appearance descriptors form an appearance
space. With properly-determined metrics, observations having a similar
appearance are identified using a nearest-neighbor-based approach. Switching
from image-space to an appearance space seems to substantially account for
occlusions, reducing their negative impact at a negligible cost in terms of
performance.

A possible alternative to appearance-based classification is the use of
template-based metrics. Such an approach uses a reference region of interest
(ROI) drawn from one or multiple frames and attempts to match it in
subsequent frames using an appropriately-constructed metric. Template-based
methods often work for partial detections, thereby accounting for occlusion
and/or noise, considering that the template needs not be perfectly or
completely matched for a successful detection to occur. An example of a
template-based method is provided by \cite{301}, which involves three CNNs,
one for template generation, one dedicated to region searching and one for
handling background areas. The method is somewhat similar to what could be
achieved by a generative adversarial network (GAN), since the ``searcher''
network attempts to fit multiple subimages within the positive detections
provided by the template component while simultaneously attempting to
maximize the distance to the negative background component. The candidate
subimages generated by the three components are fed through a loss function
which is designed to favor candidates which are closer to template regions
than to background ones. While performance-wise such a approach is claimed
to provide impressive framerates, care should be taken when using template
or reference-based methods. These are generally suited for situations where
there is no significant variation in the overall tone of the frames. Such
methods have a much higher failure rate when, for instance, the lighting
conditions change during tracking, such as when the tracked object moves
from a brightly-lit to a shaded area.

\begin{figure}[tbp]
\includegraphics[width={\textwidth}]{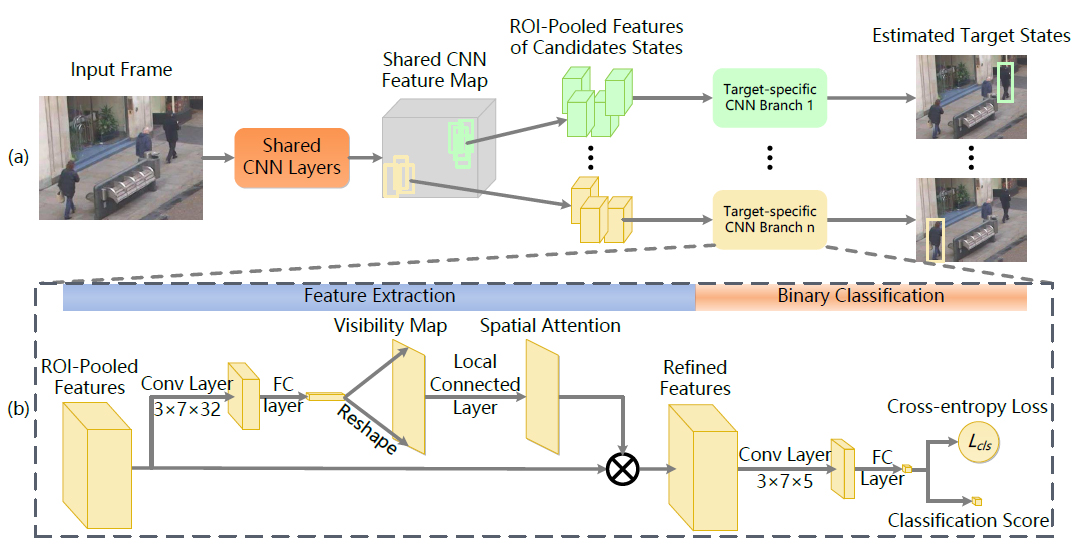}
\caption{A CNN-based model that uses ROI-pooling and shared features for
target classification \protect\cite%
{Chu_Online_Multi-Object_Tracking_ICCV_2017_paper}}
\label{fig_Chu_Online_Multi-Object_Tracking_ICCV_2017_paper}
\end{figure}

An improvement on the use of appearance and shared tracking information is
provided by \cite{Chu_Online_Multi-Object_Tracking_ICCV_2017_paper} in the
form of a CNN-based single object tracker which generates and adapts the
appearance models for multi-frame detection (Figure \ref%
{fig_Chu_Online_Multi-Object_Tracking_ICCV_2017_paper}). The use of pooling
layers and shared features accounts for drift effects caused by occlusion
and inter-object dependency, as part of a spatial and temporal attention
mechanism which is responsible for dynamically discriminating between
training candidates based on the level of occlusion. As such, training
samples are weighted based on their occlusion status, which optimizes the
training process both in terms of the resulting classification accuracy, and
performance. Generally speaking, pooling operations have two important
effects: on the one hand, the subimage of the feature map being analyzed is
increased, since a pooled feature map contains information from a larger
area of the originating image; on the other hand, the reduced size of a
pooled map means fewer computational resources are required to process it
which positively impacts performance. The major downside of pooling is that
spatial positioning is further diluted with each additional layer. Multiple
related papers involve the so called ``ROI pooling'', which commonly refers
to a pooling operation being applied to the bounding box of an identified
object in hope that the reduced representation will gain robustness to noise
and variations of the object's geometry across multiple frames. ROI pooling
is successfully used by \cite{cui18a} to improve the performance of their
CNN-based classifier. The authors observe that positioning cues are
adversely affected by pooling, to which a potential solution is to
reposition the mis-aligned ROIs via bilinear interpolation. This
reinterpretation of pooling in referred to as ``ROI align''. The gain in
performance is significant, while the authors demonstrate that the
positioning of the ROIs is stabilized.

Tracking stabilization is fundamental in automotive application, where
effects such as jittering, camera shaking and spatial/temporal noise
commonly occur. In terms of ensuring ROI stability and accuracy, occlusion
plays an important role. Some authors handle this topic extensively, such as 
\cite{Learning_to_Detect_and_Track_Visible_and_Occluded} which proposes a
deep neural network for tracking occluded body parts, by processing features
extracted from a VGG19 network. Some authors use different interpretations
of the feature concept, adapted to the specifics of autonomous driving. \cite%
{1809.10732} create custom feature maps by encoding various properties of
the detections (bounding boxes, positions, velocities, accelerations etc.)
in raster images. These images are sent though a CNN which generates raster
features that the authors demonstrate to provide more reliable correlations
and more accurate trajectories than using features derived directly from raw
data.

The idea of tracking robustness and stability is sometimes solvable using
image and object fusion. The related methods are referred to as being
``instance-aware'', meaning that a targeted object is matched across the
image space and across multiple frames by fusing identified objects with
similar characteristics. \cite{1902.08231} proposes a fusion-based method
that uses single-object tracking to identify multiple candidate instances
and subsequently builds target models for potential objects by fusing
information from detections and background cues. The models are updated
using a CNN, which ensures robustness to noise, scaling and minor variations
of the targets' appearance. As with many other related approaches, an online
implementation offloads most of the processing to an external server leaving
the embedded device from the vehicle to carry out only minor and
frequently-needed tasks. Since quick reactions of the system are crucial for
proper and safe vehicle operation, performance and a rapid response of the
underlying software is essential, which is why the online approach is
popular in this field.

Also in the context of ensuring robustness and stability, some authors apply
fusion techniques to information extracted from CNN layers. It has been
previously mentioned that important correlations can be drawn from deep and
shallow layers which can be exploited together for identifying robust
features in the data. This principle is used for instance in \cite%
{Goutam_Bhat_Unveiling_the_Power_ECCV_2018_paper}, where, in order to ensure
robustness and performance, various features extracted from layers in
different parts of a CNN are fused to form stronger characteristics which
are affected to a lesser degree by noise, spatial variation and
perturbations in the acquired images. The identified relationships between
CNN layers are exploited in order to account for lost spatial information
which occurs in deeper layers. The method is claimed to have improved
accuracy over the state-of-the-art of the time, which is consistent with the
idea of ensuring robustness and low failure rates. Deeper features are more
consistent and allow for stronger classification, while shallow features
compensate for the detrimental effects of filtering and pooling, where
relative positioning information may be lost. This allows for deep features
to be better integrated into the spatial context of the images. On a similar
note, in \cite{cvpr16_hedge_tracking} features from multiple layers which
individually constitute weak trackers are combined to form a stronger one,
by means of a hedging algorithm. The practice of using multiple weak methods
into a more effective one has significant potential and is based on the
principle that each individual weak component contains some piece of
meaningful information on the tracked object, while also having useless data
mostly found in the form of noise. By appropriately combining the
contributions of each weak component, a stronger one can be generated. As
such, methods that exploit compound classifiers typically show robustness to
variances of illumination, affine transforms, camera shaking etc. The
downside of such methods comes from the need to compute multiple groups of
weak features, which causes penalties in real-time response, while the
fusion algorithm comes with an additional overhead in terms of impacting
performance.

Alternative approaches exist which mitigate this to some extent, such as the
use of multiple sensors which directly provide data, as opposed to relying
on multiple features computed from the same camera or pair of cameras. An
example in this direction is provided in \cite{1803.10859}, where an image
gallery from a multi-camera system is fed into a CNN in an attempt to solve
multi-target multi-camera tracking and target re-identification problems.
For correct and consistent re-identification, an observation in a specific
image is matched against several ones from other cameras using correlations
as part of a similarity metric. Such correlation among images from multiple
cameras are learned during training and subsequently clustered to provide a
unified agreement between them. Eventually, after a training process that
exploits a custom triplet loss function, features are obtained to be further
used in the identification process. In terms of performance, the method
boasts substantial accuracy considering the multi-camera setup. The idea of
compositing robust features from a multi-faceted architecture is further
exploited in works such as \cite%
{ICCV17RobustObjectTrackingBasedOnTemporalAndSpatialDeepNetworks}, where a
triple-net setup is used to generate features that account for appearance,
spatial cues and temporal consistency.

\subsubsection{Ensuring Temporal Coherence}

One of the most significant challenges for autonomous driving is accounting
for temporal coherence in tracking. Since most if not all automotive
scenarios involve video and motion across multiple frames, handling image
sequence data and accounting for temporal consistency are key factors in
ensuring successful predictions, accuracy and the reliability of the systems
involved. Essentially, solving temporal tracking is a compound problem and
involves, on the one hand, tracking objects in single images considering all
the problems induced by noise, geometry and the lack of spatial information
and, on the other hand, making sure that the tracking is consistent across
multiple frames, that is, assigning correct IDs to the same objects in a
continuous video sequence.

This presents a lot of challenges, for instance when objects become occluded
in some frames and are exposed in others. In other cases, the tracked
objects suffer affine transformations across frames, of which rotation and
shearing are notoriously difficult to handle. Additionally, the objects may
change shape due to noise, aliasing and other acquisition-related artifacts
that may be present in the images, since video is rarely if ever acquired at
``high enough'' resolution and is in many cases in some lossy compressed
format. As such, the challenge is to identify features that are robust
enough to handle proper classification and to ensure temporal consistency
considering all pitfalls associated with processing video data. This often
involves a ``focus and context'' approach, where key targets are identified
in images not only by the features that they exhibit in that particular
image, but by also ensuring that the feature extraction method also accounts
for the information provided by the context which the tracked object finds
itself in. In other words, processing a key frame in a video sequence, which
provides the focus, should account for the context information that has been
drawn up from previous frames.

Where supervised algorithms are concerned, one popular approach is to
integrate recurrent components into the classifier, which inherently account
for the context provided by a set of elements from a sequence. Recurrent
neural networks (RNN) and, more specifically, long short-term memory (LSTM)
layers are frequently present in the related literature where temporal data
is concerned. When training and exploiting RNN layers to classify sequences,
the results from one frame carry over to the computations that take place
for subsequent frames. As such, when processing the current frame, resulting
detections also account for what was found in previous frames. For
automotive applications, one advantage of neural networks is that they can
be trained off-site, while the resulting model can be ported to the embedded
device in the vehicle where predictions and tracking can occur at usable
speeds. While training a recurrent network or multiple collaborating
networks can take a long time, forward-propagating new data can happen quite
fast, making these algorithms a realistic choice for real-time tracking.

LSTMs are however not the ``magic'' solution, nor the de facto method for
handling sequence data, since many authors have successfully achieved high
accuracy results using only CNNs. Additionally, many authors have found it
helpful to use dual neural networks in conjunction, where one network
processes spatial information while the other handles temporal consistency
and motion. Other methods employ siamese networks, i.e. identical
classifiers trained differently which identify different features using
similar processing. One example of a dual-streaming network is in \cite{279}
where appearance and motion are handled by a combination of CNNs which work
together within a unified framework. The motion component uses spotlight
filtering over feature maps which result from subtracting features drawn
from dual CNNs and generates a space-invariant feature map using pooling and
fusion operations. The other component handles appearance by filtering and
fusing features from a different arrangement of convolutional layers. Data
from ROIs in the acquired images is passed on to both components and motion
responses from one component are correlated with appearance responses from
the other. Both components produce feature maps which are composed together
to form space- and motion-invariant characteristics to be further used for
target identification.

Another concept which consistently appears in the related literature is
``historical matching'' where attempts are made to carry over part of the
characteristics of tracked objects across multiple frames, by building an
affinity model from shape, appearance, positional and motion cues. This is
achieved in \cite{1805.10916} using dual CNNs with multistep training, which
handle appearance matching using various filtering operations and linearly
composing the resulting features across multiple timestamps. The notion of
determining and preserving affinity is also exploited in \cite{1810.1178}
where data consisting of frame pairs several timestamps apart are fed into
dual VGG networks. The resulting features are permuted and incorporated into
association matrices which are further used to compute object affinities.
This approach has the benefit of partially accounting for occlusion using
only a limited number of frames, since the affinity of an object which is
partially occluded in one frame may be preserved if it appears fully in the
pair frame.

Ensuring the continuity of high-level features such as appearance models is
not a trivial task, and multiple solutions exist. For example \cite%
{2018111616343061} uses a CNN modified with a discriminative component
intended to correct for temporal errors that may accumulate in the
appearance of tracked objects across multiple frames. Discriminative network
behavior is also exploited in \cite{DeepTracking} where selectively trained
dual networks are used to generate and correlate appearance with a motion
stream. Also, decomposing the tracking problem into localization and motion
using multiple component networks is a frequently-encountered solution,
further exploited in works such as \cite{DLST_MM}, \cite%
{Feichtenhofer_Detect_to_Track_ICCV_2017_paper}. As such, using two networks
that work in tandem is a popular approach and seems to provide accurate
results throughout the available literature (Figure \ref%
{fig_Feichtenhofer_Detect_to_Track_ICCV_2017_paper}).

\begin{figure}[tbp]
\includegraphics[width={\textwidth}]{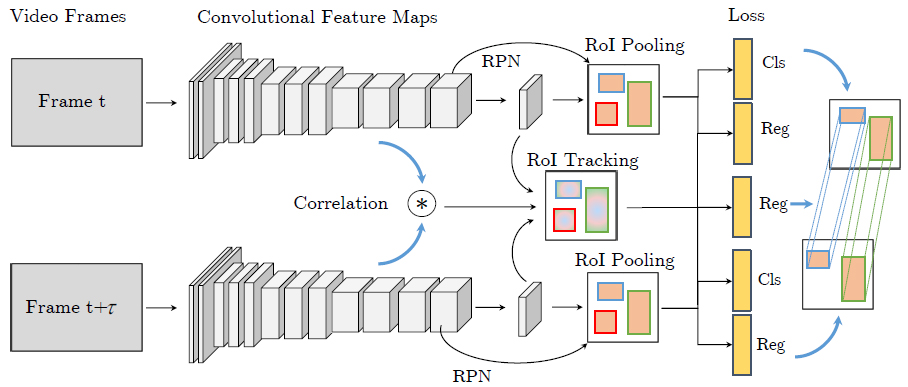}
\caption{A dual CNN detector that extracts and correlates features from
frame pairs \protect\cite{Feichtenhofer_Detect_to_Track_ICCV_2017_paper}}
\label{fig_Feichtenhofer_Detect_to_Track_ICCV_2017_paper}
\end{figure}

Some authors take this concept further by employing several such networks 
\cite{Son_Multi-Object_Tracking_With_CVPR_2017_paper}, each of which
contributes features exhibiting specific and limited correlations, which,
when joined together, from a complete appearance model of the tracked
objects. Other approaches map network components to flow graphs, the
traversal of which enables optimal cost-function and feature learning \cite%
{Schulter_Deep_Network_Flow_CVPR_2017_paper}. It is worthy of noting that
the more complicated the architecture of the classifier, the more elaborate
the training process and the poorer the performance. A careful balance
should therefore be reached between the complexity of the classifier, the
completeness of the resulting features and the amount of processing and
training data needed to produce high-accuracy results at a cost in
computational resources which is consistent with the needs of automotive
applications.

In \cite{1901.06129}, the idea of object matching from frame pairs is
further explored using a three-component setup: a siamese network
configuration handles single object tracking and generates short-term cues
in the form of tracklet images, while a modified version of GoogLeNet
generates re-identification features from multiple tracklets. The third
component is based on the idea that there may be a large overlap in the
previously-computed features, which are consequently treated as switcher
candidates. As a result, a switcher-aware logic handles the situation where
IDs of different objects may be interchanged during frame sequences mainly
as a result of partial occlusion.

It is worth mentioning that the tendency in ensuring accurate tracking is to
come up with inventive features which express increasingly-abstract
concepts. It has been demonstrated throughout the related literature that,
in general, the more abstract the feature, the more reliable it is long
term. Therefore, a lot of effort is directed toward identifying object
features that are not necessarily direct indicators of shape, position
and/or geometry, but are rather higher-level, more abstract representations
of how the object fits within the overall context of the acquired video
sequence. Examples of such concept are the previously-mentioned
``affinity''; another is ``attention'', where some authors propose
neural-network-based solutions for estimating attention and generating
attention maps. \cite{Chu_Online_Multi-Object_Tracking_ICCV_2017_paper}
computes attention features which are spatially and temporally sound using
an arrangement of ROI identification and pooling operations. \cite%
{eccv2018_mot} uses attention cues to handle the inherent noise from
conventional detection methods, as well as to compensate for frequent
interactions and overlaps among tracked targets. A two-component system
handles noise and occlusion and produces spatial attention maps by matching
similar regions from pair frames, while temporal coherence is achieved by
weighing observations across the trajectory differently, thereby assigning
them different levels of attention, which generates filtering criteria used
to successfully account for similar observations while eliminating
dissimilar ones. Another noteworthy contribution is \cite%
{7463-deep-attentive-tracking-via-reciprocative-learning}, where attention
maps are generated using reciprocative learning, where the input frame is
sent back-and-forth through several convolutional layers: in the forward
propagation phase classification scores are generated, while the
back-propagation produces attention maps from the gradients of the
previously-obtained scores. The computed maps are further used as
regularization terms within a classifier. The advantage of this approach is
its simplicity compared to other similar ones. The authors claim that their
method for generating attention features ensures long-term robustness, which
is advantageous considering that other methods that use frame pairs and no
recurrent components do not seem to work as well for very long-term
sequences.

\subsubsection{LSTM-Based Methods}

Generally, methods that are based on non-recurrent CNN-only approaches are
best suited to handle short scenes where quick reactions are required in a
brief situation that can be captured in a limited number of frames. Various
literature studies show that LSTM-based methods have more potential to
ensure the proper handling of long-term dependencies while avoiding various
mathematical pitfalls such as network parameters that end up having
extremely small values because of repeated divisions (e.g. the ``vanishing
gradient'' problem) which in practice manifests as a mis-trained network
resulting in drift effects and false positives. Handling long-term
dependencies means having to deal with occlusions to a greater extent than
in shorter term scenarios.

Most approaches combine various classifiers which handle spatial and
shape-based classification with LSTM components which account for temporal
coherence. An early example of an RNN implementation is \cite%
{aaai2017-anton-rnntracking} which uses an LSTM-based classifier to track
objects in time, across multiple frames (Figure \ref%
{fig_aaai2017-anton-rnntracking}). The authors demonstrate that an
LSTM-based approach is better suited to removing and reinserting candidate
observations to account for objects that leave/reenter the visible area of
the scene. This provides a solution to the track initiation and termination
problem based on data associations found in features obtained from the LSTM
layers. This concept is exploited further by \cite{1701.01909} where various
cues are determined to assess long-term dependencies using a dual LSTM
network. One LSTM component tracks motion, while the other handles
interactions, and the two are combined to compute similarity scores between
frames. The results show that using recurrent components to lengthy
sequences produces more reliable results than other methods which are based
on frame pairs. Some implementations using LSTM focus on
tracking-while-driving problems, which pose additional challenges compared
to most established benchmarks which use static cameras. As an alternative
to most related approaches which attempt to create models of vehicle
behavior, \cite{1704.07049} circumvent the need for vehicle modeling by
directly inputting sensor measurements into an LSTM network to predict
future vehicle positions and to analyze temporal behavior. A more elaborate
attempt is \cite{1805.05499} where instead of raw sensor data, the authors
establish several maneuver classes and feed maneuver sequences to LSTM
layers in order to generate probabilities for the occurrence of future
maneuver instances. Eventually, multiple such maneuvers can be used to
construct the trajectory and/or anticipate the intentions of the vehicles.

Furthermore, increasing the length of the sequence increases accuracy and
stability over time, up to a certain limit where the network saturates and
no longer improves. A solution to this problem would be to split the
features into multiple sub-features, followed by reconnecting them to form
more coherent long-term trajectories. This is achieved in \cite{1804.04555}
where a combined CNN and RNN-based feature extractor generates tracklets
over lengthy sequences. The tracklets are split on frames which contain
occlusions, while a recombination mechanism based on gated recurrent units
(GRUs) recombines the tracklet pieces according to their similarities,
followed by the reconstruction of the complete trajectory using polynomial
curve fitting.

\begin{figure}[tbp]
\includegraphics[width={\textwidth}]{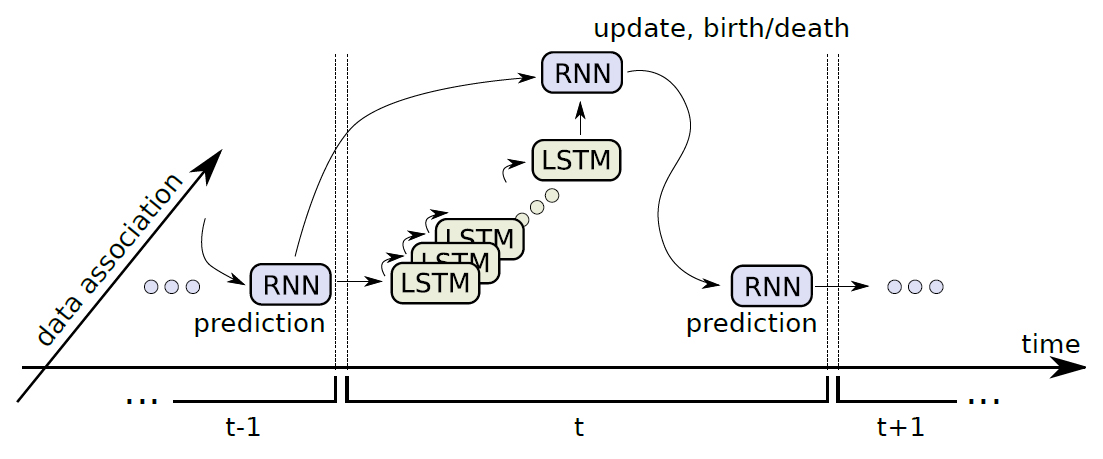}
\caption{An LSTM-based architecture used for temporal prediction 
\protect\cite{aaai2017-anton-rnntracking}}
\label{fig_aaai2017-anton-rnntracking}
\end{figure}

Some authors do further modifications to LSTM layers to produce classifiers
that generate abstract high-level features such as those found in appearance
models. A good example in this sense is \cite%
{Chanho_Kim_Multi-object_Tracking_with_ECCV_2018_paper} where LSTM layers
are modified to do multiplication operations and use customized gating
schemes between the recurrent hidden state and the derived features. The
newly-obtained LSTM layers are better at producing appearance-related
features than conventional LSTMs which excel at motion prediction. Where
trajectory estimation is concerned, LSTM-based methods exploit the gating
that takes place in the recurrent layers, as opposed to regular RNNs which
pass candidate features into the next recurrent iteration without
discriminating between them. The filters inherently present in gated LSTMs
have the potential to eliminate unwanted feature candidates which, in actual
use cases, may represent unwanted trajectory paths, while maintaining
candidates which will eventually lead to correctly-estimated motion cues.
Furthermore, LSTMs demonstrate an inherent capability to predict
trajectories that are interrupted by occlusion events or by reduced
acquisition capabilities. This idea is exploited in order to find solutions
to the problem of estimating the layout of a full environment from limited
sensor data, a concept referred to in the related literature as ``seeing
beyond seeing'' \cite{Seeing_beyond_seeing}. Given a set of sensors with
limited capability, the idea is to perform end-to-end tracking using raw
sensor data without the need to explicitly identify high-level features or
to have a pre-existing detailed model of the environment. In this sense,
recurrent architectures have the potential to predict and reconstruct
occluded parts of a particular scene from incomplete or partial raw sensor
output. The network is trained with partial data and it is updated through a
mapping mechanism that makes associations with an unoccluded scene.
Subsequently, the recurrent layers make their own internal associations and
become capable of filling in the missing gaps that the sensors have been
unable to acquire. Specifically, given a hidden state of the world which is
not directly captured by any sensor, an RNN is trained using sequences of
partial observations in an attempt to update its belief concerning the
hidden parts of the world. The resulting information is used to
``unocclude'' the scene which was initially only partially perceived through
limited sensor data. Upon training, the network is capable of defining its
own interpretation of the hidden state of the scene. The
previously-mentioned result is elaborated upon by a group which includes the
same authors \cite{2017_IJRR_Dequaire}. A similar approach previously
applied in basic robot guidance is extended for use in assisted driving. In
this case more complex information can be inferred from raw sensor input, in
the form of occupancy maps, which together with a deep network-based
architecture allow for predicting the probabilities of obstacle presence
even in occluded portions within the field of view.

\subsubsection{Miscellaneous Methods}

An interesting alternative to conventional deep learning architectures is
the use of GANs, as demonstrated in \cite{1803.03347}. GANs train generative
models and filter their results using a discriminative component. GANs are
notoriously difficult to train which is one of the reasons why they see
seldom use in the related literature. In terms of tracking, GANs alleviate
the need to compute expensive appearance features and minimize the
fragmentation that typically occurs in more conventional trajectory
prediction models. A generative component produces and updates candidate
observations of which the least updated are eliminated. The
generative-discriminative model is used in conjunction with an LSTM
component to process and classify candidate sequences. This approach has the
potential to produce high-accuracy models of human behavior, especially
group behavior. At the same time, it is significantly more lightweight than
previously-considered CNN-based solutions.

Another ``outlier'' solution in the related literature is \cite%
{Liangliang_Ren_Collaborative_Deep_Reinforcement_ECCV_2018_paper}, one of
the few efforts involving reinforcement learning for MOT applications. As
opposed to many other approaches which split the problem among different
networks and then join together the results, the authors demonstrate that
deep reinforcement learning is a one-step solution where the collaborative
interactions of multiple tracked agents are exploited in order to
simultaneously detect and track the targeted objects.

\subsubsection{Discussion}

Most of the results from the available literature focus on generating
abstract, high-level features of the observations found in the processed
images, since, generally, the more abstract the feature the more robust it
should be to transformations, noise, drift and other undesired artifacts and
effects. Most authors rely on an arrangement of CNNs where each component
has a distinct role in the system, such as learning appearance models,
geometric and spatial patterns, of learning temporal dependencies. It is
worthy of noting that a strictly CNN-based method needs substantial tweaking
and careful parameter adjustment before it can accomplish the complex task
of consistent detection in space and across multiple frames.

A system made up of multiple networks, each with its own purpose, is also
difficult to properly train, requiring lots of data and having a grater risk
of overfitting. However, complex, customized CNN solutions still seem to
provide the best accuracies within the current state-of-the-art. Most such
results also use frame pairs, or only a few elements from the video
sequence, thereby making them unreliable for long-term tracking.

LSTM-based architectures seem to show more promising results for ensuring
long-term temporal coherence, since this is what they were designed for,
while also being simpler to implement and train. For the purposes of
autonomous driving, an LSTM-based method shows promise, considering that
training should happen offline and that a heavily-optimized solution is
needed to achieve a real-time response. Designing such a system also
requires a fair amount of trial-and error since currently there is no well
established manner to predict which network architecture is suited to a
particular purpose.

There are also very few solutions based on reinforcement learning for object
tracking, especially considering that reinforcement learning has gained
substantial momentum in automotive decision making problems. Other less
popular but promising solutions, such as GAN-based predictors, may be worthy
of further study and experimentation.

One particularly promising direction for automotive tracking are solutions
that make use of limited sensor data and that are able to efficiently
predict the surrounding environment without requiring a full representation
or reconstruction of the scene. These approaches circumvent the need for
lengthy video sequences, heavy image processing and the computation of
complicated object features while being especially designed to handle
occlusion and objects outside of the immediate field of view. As such, where
automotive tracking is concerned, the available results from the
state-of-the art seem to suggest that an effective solution would make use
of partial data while being able to handle temporal correlations across
lengthy sequences using an LSTM component.

As of yet, solutions based on deep neural networks show the most promise
since they offer the most robust features while being natively designed to
solve focus-and-context problems in video sequences. In this sense, the
results which seem most promising for the complex tracking problems
described in this section are \cite{1510.07945MDnet}, \cite{1608.03773CCOT}, 
\cite{ECO1611.09224} and \cite{SANet}.

\subsection{Other Techniques}

While the current state-of-the art methods for MOT are mostly neural
network-based, there also exist a multitude of other approaches which
exploit more traditional, unsupervised means of providing reliable tracking.
Neural networks gained popularity in recent years due in no small part to
the availability of more powerful hardware, particularly GPUs, which allowed
for training models capable of handling realistic scenarios in a reasonable
amount of time. Neural networks however have the downside of needing vast
amounts of reliable training data, on the one hand, and requiring a lot of
experimentation and trial-and-error before the right design and
hyperparameter set is found for a particular scenario, on the other hand.
There are, however, situations where training data may not be readily
available, or it may not allow for sufficient generalization. Such cases
call for a more straightforward design and a more intuitive model that can
provide reliable tracking without necessarily requiring supervision. Neural
network models are harder to understand in terms of how they function, and,
while as deterministic as their non-NN counterparts, are less intuitive and
meant for use in a ``black-box'' manner. This is where other, more
transparent methods come into place.

The tracking problem can be formulated similarly to the neural-network case:
given a set of observations/appearances/segmented objects in multiple video
frames, the task is to develop a means of determining relationships among
these elements across the frames and to come up with a means of predicting
their path. Various authors formulate this problem differently, for instance
some methods involve determining tracklets in each frame and then assembling
object trajectories in a full video sequence by combining tracklets from all
or some of the frames. Traditional, non-NN-based approaches, especially
non-supervised ones, generally formulate much more straightforward models,
commonly based on a graph or flow-oriented interpretation of the tracked
scene, or on emitting hypotheses as to the potential trajectories of the
tracked targets, or otherwise formulating some probabilistic approach to
predicting the evolution of objects in time. It is worth noting that many of
the more conventional, unsupervised algorithms from the state-of-the-art do
not generalize the solution as well as a NN-based method, meaning that they
are usable in a limited number of scenarios, by comparison. Also, methods
that attempt to account for temporal consistency do not handle time
sequences as lengthy as, for instance, an LSTM network. The likely
explanation is that an unsupervised method requires far more processing
capabilities the more frame elements it is fed, unlike an NN-based method
for which, once properly trained, the amount of computational resources
required does not increase as much with the length of the associated
sequence. However, in practice, especially on an embedded device as required
in automotive tracking, porting a more conventional method may be more
convenient in terms of implementation and platform compatibility than
running a pre-trained NN model.

Another important aspect worth mentioning is that conventional methods are
much more varied in terms of their underlying algorithms, as opposed to an
NN-based architecture which features various arrangements of the same two or
three neural network types, with additional processing of layer activations
or outputs as the case may be. For this reason, we do not attempt to cover
all the approaches ever developed for object tracking, but we rather focus
on representative works featuring various successful attempts at MOT.

\subsubsection{Traditional Algorithms and Methods Focusing on
High-Performance}

The Kalman filter is a popular method with many applications in navigation
and control, particularly with regard to predicting the future path of an
object, associating multiple objects with their trajectories, while
demonstrating significant robustness to noise. Generally, Kalman-based
methods are used for simpler tracking, particularly in online scenarios
where the tracker only accesses a limited number of frames at a time,
possibly only the current and previous ones. An example of the use of the
Kalman filter is \cite{1602.00763}, where a combination of the
aforementioned filter and the Munkres algorithm as the min-cost estimator
are used in a simple setup focusing on performance. The method requires
designing a dynamic model of the tracked objects' motion, and is much more
sensitive to the type of detector employed than other approaches, however
once such parameters are well established, the simplicity of the algorithms
allows for significant real-time performance.

Similar methods are frequently used in simple scenarios where a limited
number of frames are available and the detections are accurate. In such
situations, the simplicity of the implementations allows for quick response
times even on low-spec embedded client devices. In the same spirit of
providing an easy, straightforward method that works well for simple
scenarios, \cite{1903.05625} provide an approach based on bounding-box
regression. Given multiple object bounding boxes in a sequence of frames,
the authors develop a regressor which allows the prediction of bounding box
positions in subsequent frames. This comes with some limitations,
specifically it requires that targets move only slightly from frame to
frame, and is therefore reliable in scenarios where the frame rate is high
enough and relatively stable. Furthermore, a reliable detector is a must in
such situations, and crowded scenes with frequent occlusion events are not
handled properly. As with the previous approach, this is well suited for
easy cases where robust image acquisition is available and performance and
implementation simplicity are a priority. Unfortunately, noisy images are
fairly common in automotive scenarios where, for efficiency and cost
reasons, a compromise may be made in terms of the quality and performance of
the cameras and sensors. It is often desirable that the software be robust
to noise so as to minimize the hardware costs.

In \cite{pedestrianTrajectories}, tracking is done by a particle filter for
each track. The authors use the Munkres assignment algorithm between
bounding boxes in the current input image and the previous bounding box for
each track. A cost matrix is populated with the cost for associating a
bounding box with any given previous bounding box: the Euclidean distance
between the box centers plus the size change of the box, as a bounding box
is expected to be roughly the same size in two consecutive frames. Since
boxes move and change size in bigger increments when the actors are close to
the camera, the cost is weighted by the inverse of the box size. This
approach is simple, but the assignment algorithm has an $O(n^3)$ complexity,
which is probably too high for real-time tracking.

Various attempts exist for improving noise robustness while maintaining
performance, for example in \cite{7410891}. In this case, the lifetime of
tracked objects is modeled using a Markov Decision Process (MDP). The policy
of the MDP is determined using reinforcement learning, whose objective is to
learn a similarity function for associating tracked objects. The positions
and lifetimes of the objects are modeled using transitions between MDP
states. \cite{1802.08755} also use MDPs in a more generalized scheme,
involving multiple sensors and cameras and fusing the results from multiple
MDP formulations. Note that Markov models can be limiting when it comes to
automotive tracking, since a typical scene with multiple interacting targets
does not exhibit the Markov property where the current state only depends on
the previous one. In this regard, the related literature features multiple
attempts to improve reliability. \cite{1801.06523} propose an elaborate
pipeline featuring multiview tracking, ground plane projection, maneuver
recognition and trajectory prediction using an assortment of approaches
which include Hidden Markov Models and Variational Gaussian mixture models.
Such efforts show that an improvement over traditional algorithms involves
sequencing together multiple different methods, each with its own role. As
such, there is the risk that the overall resulting approach may be too
fragmented and too cumbersome to implement, interpret and improve properly.

Works such as \cite{Maksai_Non-Markovian_Globally_Consistent_ICCV_2017_paper}
attempt to circumvent such limitations by proposing alternatives to
tried-and-tested Markov models, in this case in the form of a system which
determines behavioral patterns in an effort to ensure global consistency for
tracking results. There are multiple ways to exploit behavior in order to
guide the tracking process, for instance by learning and
minimizing/maximizing an energy function that associates behavioral patterns
to potential trajectory candidates. This concept is also exemplified by \cite%
{pami2014-anton}, who propose a method based on minimizing a continuous
energy function aimed at handling the very large space of potential
trajectory solutions, considering that a limited, discrete set of behavior
patterns impose limitations on the energy function. While such a limitation
offers better guarantees that a global optimum will eventually be reached,
it may not allow a complete representation of the system.

\begin{figure}[tbp]
\includegraphics[width={\textwidth}]{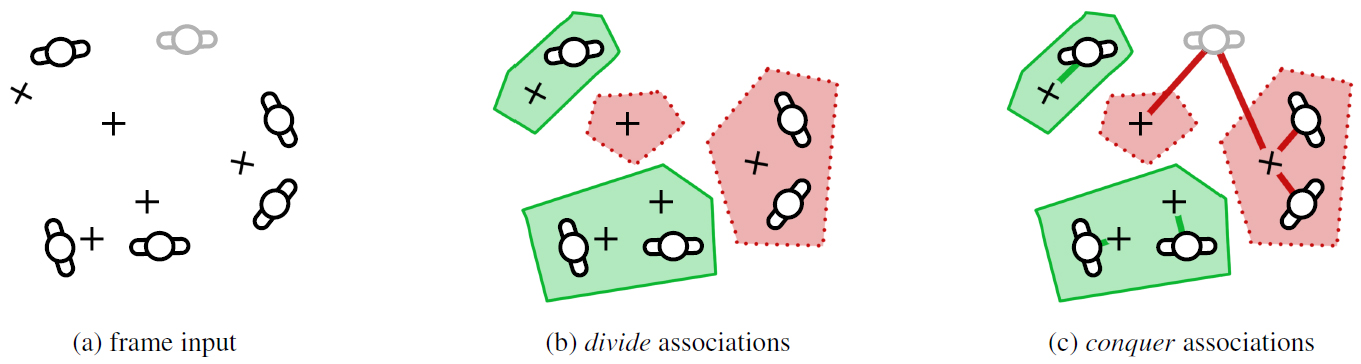}  
\caption{An example of a divide-and-conquer approach which creates
associations between detections \protect\cite{7410854}}
\label{fig_7410854}
\end{figure}

An alternative approach which is also designed to handle occlusions is \cite%
{7410854}, where the divide-and-conquer paradigm is used to partition the
solution space into smaller subsets, thereby optimizing the search for the
optimal variant. The authors note that while detections and their respective
trajectories can be extracted rather efficiently from crowded scenes, the
presence of ambiguities induced by occlusion events may raise significant
detection errors. The proposed solution involves subdividing the object
assignment problem into subproblems, followed by a selective combination of
the best features found within the subdivisions (Figure \ref{fig_7410854}).
The number and types of the features are variable, thereby accounting for
some level of flexibility for this approach. One particular downside is that
once the scene changes, the problem itself also changes and the subdivisions
need to reoccur and update, therefore making this method unsuitable for
scenes acquired from moving cameras.

A similar problem is posed in \cite%
{Bae_Robust_Online_Multi-Object_2014_CVPR_paper}, where it is also noted
that complex scenes pose tracking difficulties due to occlusion events and
similarities among different objects. This issue is handled by subdividing
object trajectories into multiple tracklets and subsequently determining a
confidence level for each such tracklet, based on its detectability and
continuity. Actual trajectories are then formed from tracklets connected
based on their confidence values. One advantage of this method in terms of
performance is that tracklets can be added to already-determined
trajectories in real-time as they become available without requiring complex
processing or additional associations. Additionally, linear discriminant
analysis is used to differentiate objects based on appearance criteria. The
concept of appearance is more extensively exploited by \cite%
{Dicle_The_Way_They_2013_ICCV_paper}, who use motion dynamics to distinguish
between targets with similar features. They approach the problem by
determining a dynamics-based similarity between tracklets using generalized
linear assignment. As such, targets are identified using motion cues, which
are complementary to more well established appearance models. While
demonstrating adequate performance and accuracy, it is worth mentioning that
motion-based features are sensitive to camera movement and are considerably
mode difficult to use in automotive situations, where motion assessment
metrics that work well for static cameras may be less reliable when the
cameras are in motion and image jittering and shaking occur.

The idea of generating appearance models using traditional means is
exemplified in \cite{MHTR_ICCV2015}, who use a combination appearance models
learned using a regularized least squares framework and a system for
generating potential solution candidates in the form of a set of track
hypotheses for each successful detection. The hypotheses are arranges in
trees, each of which are scored and selected according to the best fit in
terms of providing usable trajectories. An alternative to constructing an
elaborate appearance model is proposed by \cite{1802.09298}, who directly
involve the shape and geometry of the detections within the tracking
process, therefore using shape-based cost functions instead of ones based on
pixel clusters. Furthermore, results focusing on tracking-while-driving
problems may opt for a vehicle behavior model, or a kinematic model, as
opposed to one that is based on appearance criteria. Examples of such
approaches are \cite{7577010}, \cite{8443497}, where the authors build
models of vehicle behavior from parameters such as steering angles,
headings, offset distances, relative positions etc. Note that kinematic and
motion models are generally more suited to situations where the input
consists in data from radar, LiDAR or GPS, as opposed to image sequences. In
particular, attempting to reconstruct visual information from LiDAR point
clouds is not a trivial task and may involve elaborate reconstruction,
segmentation and registration preprocessing before a suitable detection and
tracking pipeline can be designed \cite{67061}.

Another class of results from related literature follows a different
paradigm. Instead of employing complex energy minimization functions and/or
statistical modeling, other authors opt for a simpler, faster approach that
works with a limited amount of information drawn from the video frames. The
motivation is that in some cases the scenarios may be simple enough that a
straightforward method that alleviates the need for extended processing may
prove just as effective as more complex and elaborate counterparts. An
example in this direction is \cite{1517Bochinski20176} whose method is based
on scoring detections by determining overlaps between their bounding boxes
across multiple consecutive frames. A scoring system is then developed based
on these overlaps and, depending on the resulting scores, trajectories are
formed from sets of successive overlaps of the same bounding boxes. Such a
method does not directly handle crowded scenes, occlusions or fast moving
objects whose positions are far apart in consecutive frames, however it may
present a suitable compromise in terms of accuracy in scenarios where
performance is detrimental and the embedded hardware may not allow for more
complex processing. An additional important consideration for this type of
problem is how the tracking method is evaluated.

Most authors use a common, established set of benchmarks which, while having
a certain degree of generality, cannot cover every situation that a vehicle
might be found in. As such, some authors such as \cite{ristani2016MTMC}
devote their work to developing performance and evaluation metrics and data
sets which allow for covering a wide range of potential problems which may
arise in MOT scenarios. As such, the choice in the method used for tracking
is as much a consequence of the diversity of situations and events claimed
to be covered by the method, as it results from the evaluation performed by
the authors. For example, as was the case for NN-based methods, most
evaluations are done for scenes with static cameras, which are only partly
relevant for automotive applications. The advantage of the methods presented
thus far lies in the fact that they generally outperform their counterparts
in terms of the required processing power and computational resources, which
is a plus for vehicle-based tracking where the client device is usually a
low-power solution. Furthermore, some methods can be extended rather easily,
as the need may be, for instance by incorporating additional features or
criteria when assembling trajectories from individual detections, by finding
an optimizer that ensures additional robustness, or, as is already the case
with some of the previously-mentioned papers, by incorporating a
light-weight supervised classifier in order to boost detection and tracking
accuracy.

\subsubsection{Methods Based on Graphs and Flow Models}

A significant number of results from the related literature present the
tracking solution as a graph search problem or otherwise model the tracking
scene using a dependency graph or flow model. There are multiple advantages
to using such an approach: graph-based models tailor well to the
multitracking problem since, like a graph, it is formed from inter-related
nodes each with a distinct set of parameter values. The relationships that
can be determined among tracked objects or a set of trajectory candidates
can be modeled using edges with edge costs. Graph theory is well understood
and graph traversal and search algorithms can be widely found, with
implementations readily available on most platforms. Likewise, flow models
can be seen as an alternative interpretation of graphs, with node
dependencies modeled through operators and dependency functions, forming an
interconnected system. Unlike a traditional graph, data from a flow model
progresses in an established direction which starts from initial components
where acquired data is handled as input; the data then traverses
intermediate nodes where it is processed in some manner and ends up at
terminal nodes where the results are obtained and exploited. Like graphs,
flow models allow for loops which implement refinement techniques and
in-depth processing via multiple local iterations.

Most methods which exploit graphs and flow models attempt to solve the
tracking problem using a minimum path or minimum cost - type approach. An
example in this sense is \cite{1408.3304}, where multi-object tracking is
modeled using a network flow model subjected to min-cost optimization. Each
path through the flow model represents a potential trajectory, formed by
concatenating individual detections from each frame. Occlusion events are
modeled as multiple potential directions arising from the occlusion node and
the proposed solution handles the resulting ambiguities by incorporating
pairwise costs into the flow network.

A more straightforward solution is presented by \cite{BerclazFTF11}, who
solve multi-tracking using dynamic programming and formulate the scenario as
a linear program. They subsequently handle the large number of resulting
variables and constraints using k-shortest paths. One advantage of this
method seems to be that it allows for reliable tracking from only four
overlapping low resolution low fps video streams, which is in line with the
cost-effectiveness required by automotive applications.

Another related solution is \cite{greedy_fahim_albert}, where a cost
function is developed from estimating the number of potential trajectories
as well as their origins and end frames. Then, the scenario is handled as a
shortest-path problem in a graph which the authors solve using a greedy
algorithm. This approach has the advantage that it uses well-established
methods, therefore affording some level of simplicity to understanding and
implementing the algorithms.

In \cite{ristaniAccv14}, a similar graph-based solution divides the problem
into multiple subproblems by exploring several graph partitioning mechanisms
and uses greedy search based on Adaptive Label Iterative Conditional Modes.
Partitioning allows for successful disassociation of object identities in
circumstances where said identities might be confused with one another.
Also, methods based on solution space partitioning have the advantage of
being highly scalable, therefore allowing fine tuning of their parameters in
order to achieve a trade-off between accuracy and performance. Multiple
extensions of the graph-based problem exists in the related literature, for
instance when multiple other criteria are incorporated into the search
method. \cite{GMCP-Tracker_ECCV12} incorporate appearance and motion-based
cues into their data association mechanism, which is modeled using a global
graph representation and makes use of Generalized Minimum Clique Graphs to
locate representative tracklets in each frame. Among other advantages, this
allows for a longer time span to be handled, albeit for each object
individually.

Another related approach is provided in \cite{CVIU2016}, where the solution
consists in a collaborative model which makes use of a detector and multiple
individual trackers, whose interdependencies are determined by finding
associations with key samples from each detected region in the processed
frames. These interdependencies are further exploited via a sample selection
method to generate and update appearance models for each tracker.

\begin{figure}[tbp]
\centering
\includegraphics[width={0.8\textwidth}]{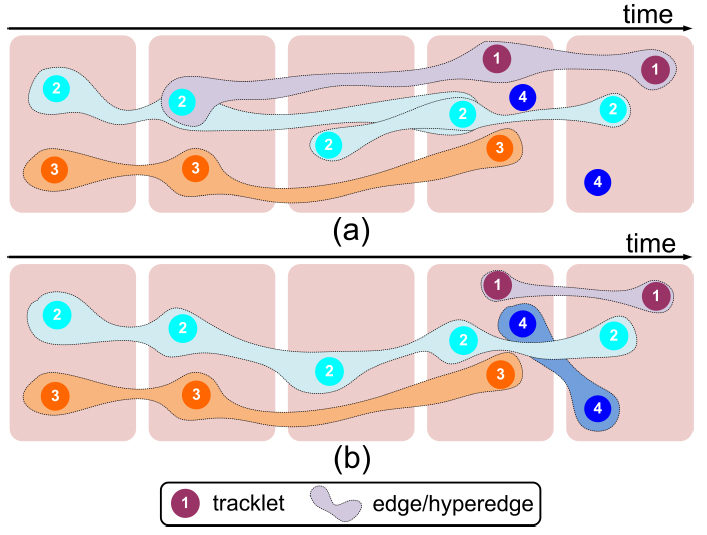}  
\caption{Generation of trajectories by determining higher order dependencies
between tracklets via a hypergraph model with edge shapes determined using a
learning method \protect\cite{aaai19a}}
\label{fig_aaai19a}
\end{figure}

As extensions of the more traditional graph-based models which use greedy
algorithms to search for suitable candidate solutions and update the
resulting models in subsequent processing steps, some authors handle the
problem using hypergraphs. These extend the concept of classical graphs by
generalizing the role of graph edges. In a conventional graph an edge joins
two nodes, while in a hypergraph edges are sets of arbitrary combinations of
nodes. Therefore an edge in a hypergraph connects to multiple nodes, instead
of just two as in the traditional case. This structure has the potential to
form more extensive and complete models using a singular unified concept and
to alleviate the need for costly solution space partitioning or subdivision
mechanisms. Another use of the hypergraph concept is provided by \cite%
{CVPR2014_HyperGraphMultiTargetsTracker}, who build a hypergraph-based model
to generate meaningful data associations capable of handling the problem of
targets with similar appearance and in close proximity to one-another, a
situation frequently encountered in crowded scenes. The hypergraph model
allows for the formulation of higher-order relationships among various
detections, which, as mentioned in previous sections, have the potential to
ensure robustness against simple transformations, noise and various other
spatial and temporal inaccuracies. The method is based on grouping dense
neighborhoods of tracklets hierarchically, forming multiple layers which
enable more fine-grained descriptions of the relationships that exists in
each such neighborhood. A related but much more recent result \cite{aaai19a}
is also based on the notion that hypergraphs allow for determining higher
order dependencies among tracklets, but in this case the parameters of the
hypergraph edges are learned using an SSVM (structural support vector
machine), as opposed to being determined empirically. Trajectories are
established as a result of determining higher order dependencies by
rearranging the edges of the hypergraph so as to conform to several
constraints and affinity criteria. While demonstrating robustness to affine
transforms and noise, such methods still cannot handle complex crowded
scenes with multiple occlusions and, compared to previously-mentioned
methods, suffer some penalties in terms of performance, since updating the
various parameters of hypergraph edges can be computationally costly.

\section{Trajectory Prediction Methods}

Autonomous cars need to have the ability to predict the future motion of
surrounding vehicles in order to navigate through complex traffic scenarios
safely and efficiently. The existence of multiple interacting agents, the
multi-modal nature of driver behavior, and the inherent uncertainty involved
make motion prediction a challenging problem. An autonomous vehicle deployed
in complex traffic needs to balance two factors: the safety of humans in and
around it, and efficient motion without stalling traffic. The vehicle should
also take the initiative, such as deciding when to change lanes, cross
unsignalized intersections, or overtake other vehicles \cite{1805.05499}.
This requires the autonomous car to have some ability to reason about the
future state of the environment.

Other difficulties come from that requirements that such a system must be
sensitive to exceptional, rarely happening situations. It should not only
consider physical quantities but also information about the drivers’
intentions and, because of the great number of possibilities involved, it
should take into account only a reasonable subset of possible future scene
evolutions \cite{Integrated_Maneuver-Based_Trajectory}.

One way to plan a safe maneuver is to understand the intent of other traffic
participants, i.e. the combination of discrete high-level behaviors as well
as the continuous trajectories describing future motion \cite%
{intentnet_corl18}. Predicting other traffic participants trajectories is a
crucial task for an autonomous vehicle, in order to avoid collisions on its
planned trajectory. Even if trajectory prediction is not a deterministic
task, it is possible to specify the most likely trajectory \cite%
{IROS13_PIN_161867_}.

Certain considerations about vehicle dynamics can provide partial knowledge
on the future. For instance, a vehicle moving at a given speed needs a
certain time to fully stop and the curvature of its trajectory has to be
under a certain value in order to keep stability. On the other hand, even if
each driver has its own habits, it is possible to identify some common
driving maneuvers based on traffic rules, or to assume that drivers keep
some level of comfort while driving \cite{IROS13_PIN_161867_}. In order to
effectively and safely interact with humans, trajectory prediction needs to
be both precise and computationally efficient \cite%
{Nikhil_Convolutional_Neural_Network_for_Trajectory_Prediction_ECCVW_2018_paper}%
.

A recent white paper \cite{safety-first-for-automated-driving} states that a
solution for the prediction and planning tasks of an autonomous car may
consider a combination of the following properties:

\begin{itemize}
\item \textit{Predicting only a short time into the future.} The likelihood
of an accurate prediction is indirectly related to the time between the
current state and the point in time it refers to, i.e. the further the
predicted state is in the future, the less likely it is that the prediction
is correct;

\item \textit{Relying on physics where possible}, using dynamic models of
road users that form the basis of motion prediction. A classification of
relevant objects is a necessary input to be able to discriminate between
various models;

\item \textit{Considering the compliance of other road users with traffic
rules to a valid extent.} For example, the ego car should cross
intersections with green traffic lights without stopping, relying on other
road users to follow the rule of stopping at red lights. In addition to
this, foreseeable non-compliant behavior to traffic rules, e.g. pedestrians
crossing red lights in urban areas, needs to be taken into account,
supported by defensive drive planning;

\item \textit{Predicting the situation} to further increase the likelihood
of road user prediction being correct. For example, the future behavior of
other road users when driving in a traffic jam differs greatly to their
behavior in flowing traffic.
\end{itemize}

Further, it asserts that the interpretation and prediction system should
understand not only the worst-case behavior of other road users (possible
vulnerable ones, i.e. who may not obey all traffic rules), but their
worst-case reasonable behavior. This allows it to make reasonable and
physically possible assumptions about other road users. The automated
driving system should make a naturalistic assumption, just as humans do,
about the reasonable behavior of others. These assumptions need to be
adaptable to local requirements so that they meet locally different
``driving cultures''.

\subsection{Problem Description}

\begin{figure}[tbp]
\centering
\includegraphics[width={12 cm}]{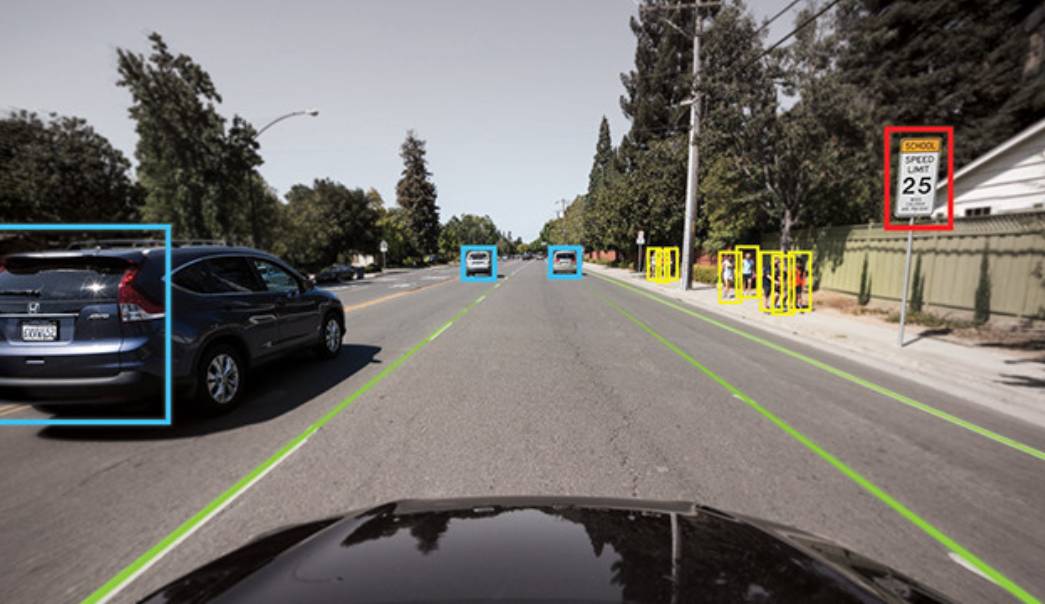}
\caption{Detected objects: cars (blue), pedestrians (yellow), road lanes
(green) \protect\cite{FULLTEXT01_mp}}
\label{fig_FULLTEXT01_mp}
\end{figure}

To tackle the trajectory prediction task, one should assume to have access
to real-time data streams coming from sensors such as lidar, radar or
camera, installed aboard the self-driving vehicle and that there already
exists a functioning tracking system that allows detection and tracking of
traffic actors in real-time. Examples of pieces of information that describe
an actor are: bounding box, position, velocity, acceleration, heading, and
heading change rate. It may also be needed to have mapping data of the area
where the ego car is driving, i.e. road and crosswalk locations, lane
directions, and other relevant map information. Past and future positions
are represented in an ego car-centric coordinate system. Also, one needs to
model the static context with road and crosswalk polygons, as well as lane
directions and boundaries: road polygons describe drivable surface, lanes
describe the driving path, and crosswalk polygons describe the road surface
used for pedestrian crossing \cite{1808.05819}. An example of available
information on which the prediction module can operate is presented in
Figure \ref{fig_FULLTEXT01_mp}.

More formally, considering the future as a consequence of a series of past
events, a \textit{prediction} entails reasoning about probable outcomes
based on past observations \cite{DESIRE}. Let $X_{t}^{i}$ be a vector with
the spatial coordinates of actor $i$ at observation time $t$, with $t\in
\{1,2,...,T_{obs}\}$, where $T_{obs}$ is the present time step in the series
of observations. The past trajectory of actor $i$ is a sequence $%
X^{i}=\{X_{1}^{i},X_{2}^{i},...,X_{T_{obs}}^{i}\}$. Based on the past
trajectories of \textit{all actors}, one needs to estimate the future
trajectories of all actors, i.e. $\hat{Y}^{i}=\{\hat{Y}_{T_{obs}+1}^{i},\hat{%
Y}_{T_{obs}+2}^{i},...,\hat{Y}_{T_{pred}}^{i}\}$.

It is also possible to first generate the trajectories in the Frenet frame
along the current lane of the vehicle, then convert it to the initial
Cartesian coordinate system \cite{IROS13_PIN_161867_}. The Frenet coordinate
system is useful to simplify the motion equations when cars travel on curved
roads. It consists of longitudinal and lateral axes, denoted as $s$ and $d$,
respectively. The curve that goes through the center of the road determines
the $s$ axis and indicates how far along the car is on the road. The $d$
axis indicates the lateral displacement of the car. $d$ is 0 on the center
of the road and its absolute value increases with the distance from the
center. Also, it can be positive or negative, depending on the side of the
road.

\subsection{Classification of Methods}

There are several classification approaches presented in the literature
regarding trajectory planning methods.

An online tutorial \cite{Prediction_in_Autonomous_Vehicle_-_TDS}
distinguishes the following categories:

\begin{enumerate}
\item \textbf{Model-based approaches.} They identify common behaviors of the
vehicle, e.g. changing lane, turning left, turning right, determining
maximum turning speed, etc. A model is created for each possible trajectory
the vehicle can go and then probabilities are computed for all these models.
One of the simplest approaches to compute the probabilities is the
autonomous multiple modal (AMM) algorithm. First, the states of the vehicle
at times $t-1 $ and $t$ are observed. Then the process model is computed at
time $t-1$ resulting in the expected states for time $t$. Then the
likelihood of the expected state with the observed state is compared, and
the probability of the model at time $t$ is computed. Finally, the model
with the highest probability is selected;

\item \textbf{Data-driven approaches.} In these approaches a black box model
(usually a neural network) is trained using a large quantity of training
data. After training, the model will be applied to the observed behavior in
order to provide the prediction. The training of the model is usually
computationally expensive and is made offline. On the other hand, the
prediction of the trajectories, once the model is trained, is quite fast and
can be made online, i.e. in real-time. Some of these methods also employ
unsupervised clustering of trajectories using e.g. spectral clustering or
agglomerative clustering, and define a trajectory pattern for each cluster.
In the prediction stage, the vehicle partial trajectory is observed, it is
compared with the prototype trajectories, and then the trajectory most
similar to a prototype is predicted.
\end{enumerate}

A survey \cite{s40648-014-0001-z} proposes a different classification based
on three increasingly abstract levels, summarized in Figure \ref%
{fig_s40648-014-0001-z}.

\begin{figure}[tbp]
\centering
\includegraphics[width={\textwidth}]{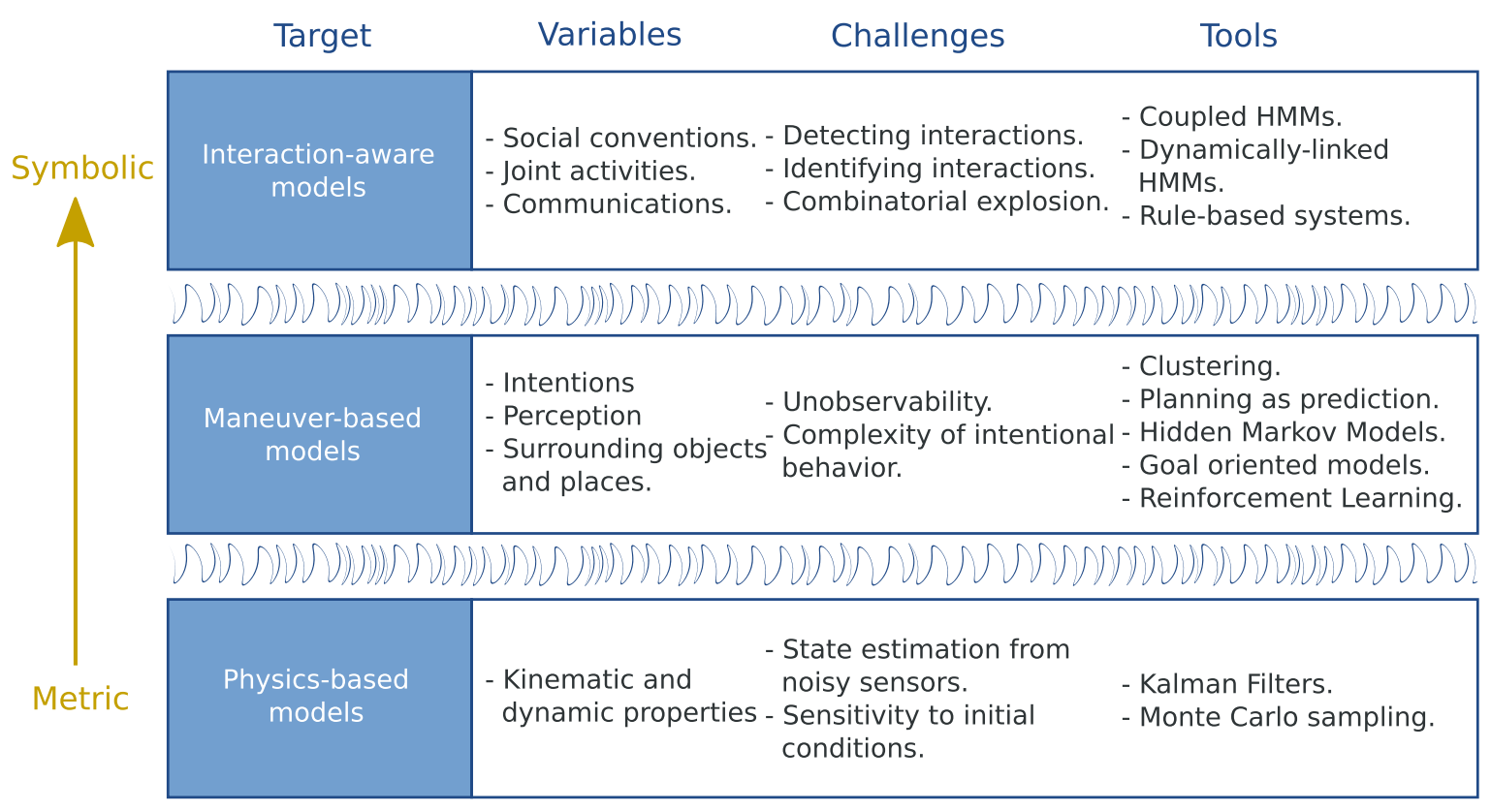}
\caption{Classification of motion models \protect\cite{s40648-014-0001-z}}
\label{fig_s40648-014-0001-z}
\end{figure}

\begin{enumerate}
\item \textbf{Physics-based motion models.} They represent vehicles as
dynamic entities governed by the laws of physics. Future motion is predicted
using dynamic and kinematic models linking some control inputs (e.g.
steering, acceleration), car properties (e.g. weight) and external
conditions (e.g. friction coefficient of the road surface) to the evolution
of the state of the vehicle (e.g. position, heading, speed). \textit{%
Advantages.} Such models are very often used for trajectory prediction and
collision risk estimation in the context of road safety. They are more or
less complex depending on how fine-grained the representation of the
dynamics and kinematics of the vehicle is, how uncertainties are handled,
whether or not the geometry of the road is taken into account, etc. \textit{%
Disadvantages.} Since they only rely on the low level properties of motion,
physics-based motion models are limited to short-term (e.g., less than a
second) motion prediction. Typically, they are unable to anticipate any
change in the motion of the car caused by the execution of a particular
maneuver (e.g., slowing down, turning at constant speed, then accelerating
to make a turn at an intersection) or changes caused by external factors
(e.g., slowing down because of a vehicle in front);

\item \textbf{Maneuver-based motion models.} They represent vehicles as
independent maneuvering entities, i.e. they assume that the motion of a
vehicle on the road network corresponds to a series of maneuvers executed
independently from the other vehicles. Trajectory prediction is based on the
early recognition of the maneuvers that drivers intend to perform. If one
can identify the maneuver intention of a driver, one can assume that the
future motion of the vehicle will match that maneuver. \textit{Advantages.}
Because of the a priori information, the derived trajectories are more
relevant and reliable in the long term than the ones derived from
physics-based motion models. Maneuver-based motion models are based either
on prototype trajectories or on maneuver intention estimation. \textit{%
Disadvantages.} In practice, the assumption that vehicles move independently
from each other does not hold. Vehicles share the road with others, and the
maneuvers performed by one vehicle necessarily influences the maneuvers of
others. Inter-vehicle dependencies are particularly strong at road
intersections, where priority rules force vehicles to take into account the
maneuvers performed by the others. Disregarding these dependencies can lead
to erroneous interpretations of the situations and to poor evaluations of
the risk;

\item \textbf{Interaction-aware motion models. }They represent vehicles as
maneuvering entities which interact with one another, i.e. the motion of a
vehicle is assumed to be influenced by the motion of the other vehicles in
the scene. \textit{Advantages.} Taking into account the dependencies between
the vehicles leads to a better interpretation of their motion compared to
the maneuver-based motion models. As a result, they contribute to a better
understanding of the situation and a more reliable evaluation of the risk.
They are based either on prototype trajectories or on dynamic Bayesian
networks. The interaction-aware motion models are the most comprehensive
models proposed so far. They allow longer-term predictions compared to
physics-based motion models, and are more reliable than maneuver-based
motion models since they account for the dependencies between the vehicles. 
\textit{Disadvantages.} Computing all the potential trajectories of the
vehicles exhaustively is computationally expensive and may not be compatible
with real-time usage.
\end{enumerate}

A classification somewhat similar with the previous two is mentioned in \cite%
{1216567}, which distinguishes the following motion prediction categories of
methods:

\begin{enumerate}
\item \textbf{Learning-based motion prediction}: learning from the
observation of the past movements of vehicles in order to predict the future
motion;

\item \textbf{Model-based motion prediction}: using motion models;

\item \textbf{Motion prediction with a cognitive architecture}: trying to
reproduce human behavior.
\end{enumerate}

Overall, the main difficulty faced by these approaches is that in order to
reliably estimate the risk of a traffic situation it is necessary to reason
at a high level about a set of interacting maneuvering entities, taking into
account uncertainties associated with the data and the models. This
high-level reasoning is computationally expensive, and not always compatible
with real-time risk estimation. For this reason, a lot of effort has been
put recently into designing novel, more efficient risk estimation algorithms
which do not need to predict all the possible future trajectories of all the
vehicles in the scene and check for collisions. Instead, algorithms have
been proposed which focus on the most relevant trajectories to speed up the
computation, or to use alternative risk indicators such as conflicts between
maneuver intentions. The choice of a risk assessment method is tightly
coupled with the choice of a motion model. Therefore, the authors of \cite%
{s40648-014-0001-z} believe that major improvements in this field will be
brought by approaches which jointly address vehicle motion modeling and risk
estimation.

In the rest of this section, we present some specific approaches classified
by their main prediction ``paradigm'', namely neural networks and other
methods, most of which use some kind of stochastic representation of the
actors' behavior in the environment. This is especially useful since some
works use the same model to address different abstraction levels of the
trajectory prediction task.

\subsection{Methods Using Neural Networks}

Many of the approaches presented in the literature that are based on neural
networks use either recurrent neural network (RNNs) which explicitly take
into account a history composed of the past states of the actors, or simpler
convolutional neural networks (CNNs).

One of the most interesting systems, albeit quite complex, is DESIRE \cite%
{DESIRE}, which has the goal of predicting the future locations of multiple
interacting agents in dynamic (driving) scenes. It considers the multi-modal
nature of the future prediction, i.e. given the same context, the future may
vary. It may foresee the potential future outcomes and make a strategic
prediction based on that, and it can reason not only from the past motion
history, but also from the scene context as well as the interactions among
the agents. DESIRE achieves these goals in a single end-to-end trainable
neural network model, while being computationally efficient. Using a deep
learning framework, DESIRE can simultaneously: generate diverse hypotheses
to reflect a distribution over plausible futures, reason about the
interactions between multiple dynamic objects and the scene context, and
rank and refine hypotheses with consideration of long-term future rewards.

The corresponding optimization problem tries to maximize the potential
future reward of the prediction, using the following mechanisms (Figure \ref%
{fig_DESIRE}):

\begin{figure}[tbp]
\centering
\includegraphics[width={\textwidth}]{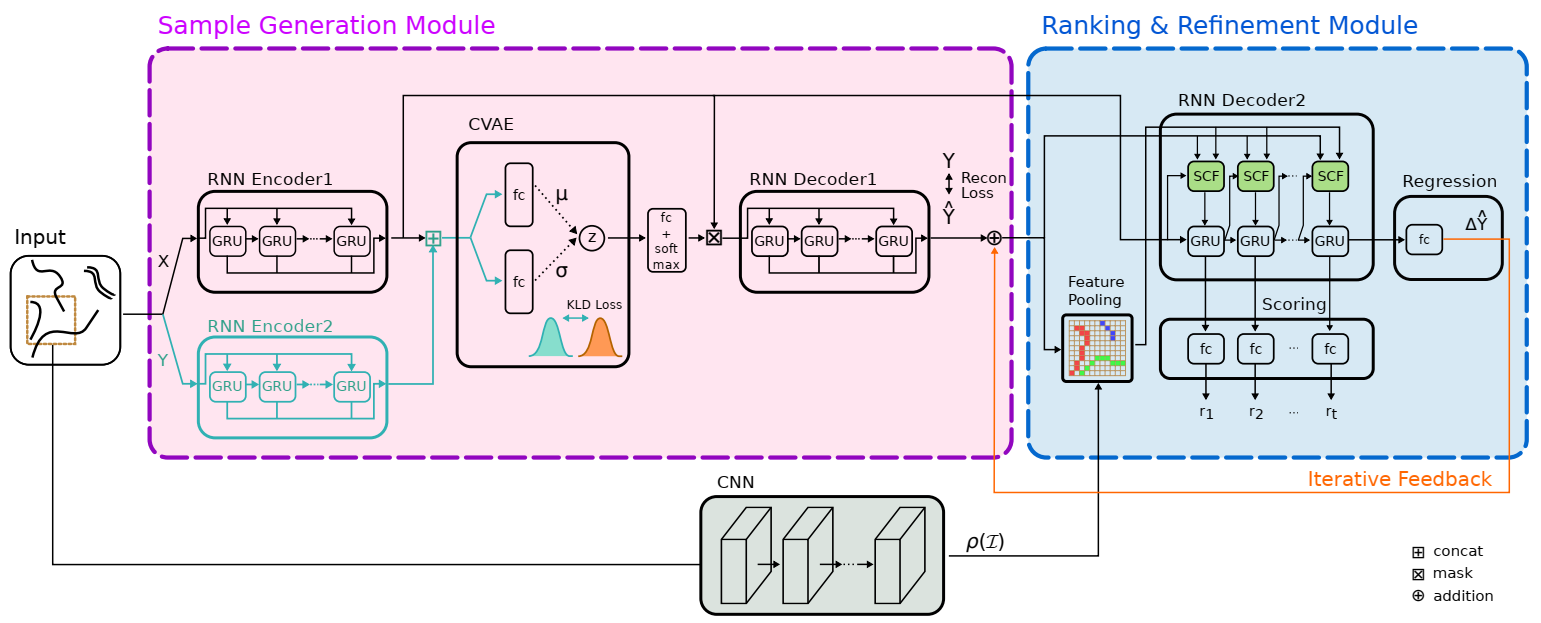}
\caption{The DESIRE architecture. \textit{GRU means Gated Recurrent Unit and
SCF means Scene Context Fusion unit. The aquamarine paths are used only
during training} \protect\cite{DESIRE}}
\label{fig_DESIRE}
\end{figure}

\begin{enumerate}
\item \textit{Diverse sample generation:} a conditional variational
auto-encoder (CVAE) is used to learn a sampling model that, given
observations of past trajectories, produces a diverse set of prediction
hypotheses to capture the multimodality of the space of plausible futures.
The CVAE introduces a latent variable to account for the ambiguity of the
future, which is combined with an RNN that encodes the past trajectories, to
generate hypotheses using another RNN. Essentially, a CVAE introduces
stochastic latent variables $z_i$ that are learned to encode a diverse set
of predictions $\mathbf{Y}_t$ given input $\mathbf{X}_t$, making it suitable
for modeling one-to-many mappings;

\item \textit{IOC-based ranking and refinement:} a ranking module determines
the most likely hypotheses, while incorporating scene context and
interactions. Since an optimal policy is hard to determine where multiple
agents make strategic interdependent choices, the ranking objective is
formulated to account for potential future rewards similar to inverse
optimal control (IOC) or inverse reinforcement learning (IRL). This also
ensures generalization to new situations further into the future, given
limited training data. The module is trained in a multitask framework with a
regression-based refinement of the predicted samples. In the testing phase,
there are multiple iterations in order to obtain more accurate refinements
of the future prediction. Predicting a distant future can be far more
challenging than predicting a closer one. Therefore, an agent is trained to
choose its actions that maximizes long-term rewards to achieve its goal.
Instead of designing a reward function manually, IOC learns an unknown
reward function. The RNN model assigns rewards to each prediction hypothesis
and measures its goodness based on the accumulated long-term rewards;

\item \textit{Scene context fusion:} this module aggregates the interactions
between agents and the scene context encoded by a CNN. The fused embedding
is channeled to the RNN scoring module and allows to produce the rewards
based on the contextual information.
\end{enumerate}

In \cite{E371Final}, a method to predict trajectories of surrounding
vehicles is proposed using a long short-term memory (LSTM) network, with the
goal of taking into account the relationship between the ego car and
surrounding vehicles.

The LSTM is a type of recurrent neural network (RNN) capable of learning
long-term dependencies. Generally, an RNN has a vanishing gradient problem.
An LSTM is able to deal with this through a forget gate, designed to control
the information between the memory cells in order to store the most relevant
previous data.

The proposed method considers the ego car and four surrounding vehicles. It
is assumed that drivers generally pay attention to the relative distance and
speed with respect to the other cars when they intend to change a lane.
Based on this assumption, the relative amounts between the target and the
four surrounding vehicles are used as the input of the LSTM network. The
feature vector $\mathbf{x}_t$ at time step $t$ is defined by twelve
features: lateral position of target vehicle, longitudinal position of
target vehicle, lateral speed of target vehicle, longitudinal speed of
target vehicle, relative distance between target and preceding vehicle,
relative speed between target and preceding vehicle, relative distance
between target and following vehicle, relative speed between target and
following vehicle, relative distance between target and lead vehicle,
relative speed between target and lead vehicle, relative distance between
target and ego vehicle, and relative speed between target and ego vehicle.
The input vector of the LSTM network is a sequence data with $\mathbf{x}_t $%
's for past time steps. The output is the feature vector at the next time
step $t + 1$. A trajectory is predicted by iteratively using the output
result of the network as the input vector for the subsequent time step.

In \cite{1704.07049} an efficient trajectory prediction framework is
proposed, which is also based on an LSTM. This approach is data-driven and
learns complex behaviors of the vehicles from a massive amount of trajectory
data. The LSTM receives the coordinates and velocities of the surrounding
vehicles as inputs and produces probabilistic information about the future
location of the vehicles over an occupancy grid map (Figure \ref%
{fig_1704.07049_1}). The experiments show that the proposed method has
better prediction accuracy than Kalman filtering.

\begin{figure}[tbp]
\centering
\includegraphics[width={12 cm}]{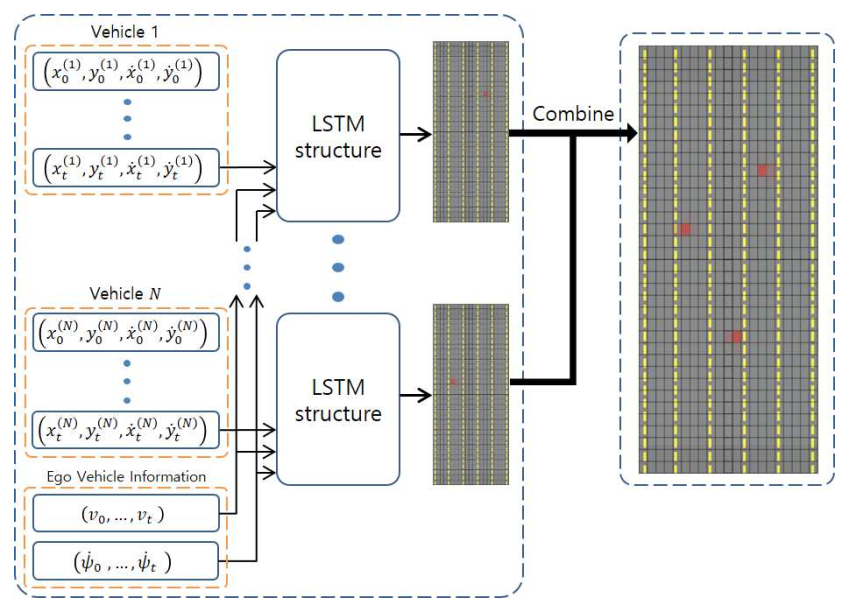}
\caption{The architecture of the system \protect\cite{1704.07049}}
\label{fig_1704.07049_1}
\end{figure}

\begin{figure}[tbp]
\centering
\includegraphics[width={10 cm}]{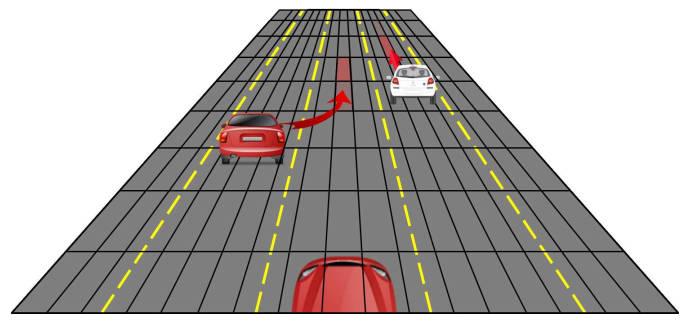}
\caption{An example of an occupancy grid map \protect\cite{1704.07049}}
\label{fig_1704.07049_2}
\end{figure}

The occupancy grid map is widely adopted for probabilistic localization and
mapping. It reflects the uncertainty of the predicted trajectories. In \cite%
{1704.07049}, the occupancy grid map is constructed by partitioning the
range under consideration into several grid cells. The grid size is
determined such that a grid cell approximately covers the quarter lane to
recognize the movement of the vehicle on same lane as well as length of the
vehicle (Figure \ref{fig_1704.07049_2}).

When predictions are needed for different time ranges (e.g., $\Delta$ =
0.5s, 1s, 2s), the LSTM is trained independently for each time range. The
LSTM produces the probability of occupancy for each grid cell. Let $(x, y)$
be a two dimensional index for the occupancy grid. Then the softmax layer in
the $i$\textsuperscript{th} LSTM produces the probability $P_o (i_x ,i_y )$
for the grid element $(i_x ,i_y )$. Finally, the outputs of the $n$ LSTMs
are combined using $P_o (i_x ,i_y ) = 1 - \prod\limits_{i = 1}^n {\left( {1
- P_o^{(i)} (i_x ,i_y )} \right)} $. The probability of occupancy $P_o (i_x
,i_y )$ summarizes the prediction of the future trajectory for all $n$
vehicles in the single map.

Alternatively, the same LSTM architecture can be used to directly predict
the coordinates of a vehicle as a regression task. Instead of using the
softmax layer to compute probabilities, the system can produce two real
coordinate values $x$ and $y$.

In \cite{1805.05499}, another LSTM model is described for interaction-aware
motion prediction. Confidence values are assigned to the maneuvers that are
performed by vehicles. Based on them, a multi-modal distribution over future
motions is computed. More specifically, the model assigns probabilities for
different maneuver classes, and outputs maneuver specific predictions for
each maneuver class. The LSTM uses as input the track histories of the ego
vehicle and its surrounding vehicles, and the lane structure of the freeway.
It assigns confidence values to six maneuver classes and predicts a
multi-modal distribution of the possibilities of future motion.

Taking into account the time constraints of a real-time system, \cite%
{1808.05819} uses simple feed-forward CNN architectures for the prediction
task. Instead of manually defining features that represent the context for
each actor, the authors rasterize the scene for each actor into an RGB
image. Then, they train the CNN using these rasterized images as inputs to
predict the actors' trajectories, where the network automatically infers the
relevant features. Optionally, the model can also take as input a current
state of the actor represented as a vector containing velocity,
acceleration, and heading change rate (position and heading are not required
because they are implicitly included in the raster image), and concatenate
the resulting vector with the flattened output of the base CNN. Finally, the
combined features are passed through a fully connected layer.

A similar approach is used in \cite{1809.10732}, which presents a method to
predict multiple possible trajectories of actors while also estimating their
probabilities. It encodes each actor's surrounding context into a raster
image, used as input by a deep convolutional network to automatically derive
the relevant features for the task. Given the raster image and the state
estimates of actors at a time step, the CNN is used to predict a multitude
of possible future state sequences, as well as the probability of each
sequence.

As part of a complete software stack for autonomous driving, NVIDIA created
a system based on a CNN, called PilotNet \cite{1704.07911}, which outputs
steering angles given images of the road ahead. This system is trained using
road images paired with the steering angles generated by a human driving a
car that collects data. The authors developed a method for determining which
elements in the road image influence its steering decision the most. It
seems that in addition to learning the obvious features such as lane
markings, edges of roads and other cars, the system learns more subtle
features that would be hard to anticipate and program by engineers, e.g.,
bushes lining the edge of the road and atypical vehicle classes, while
ignoring structures in the camera images that are not relevant to driving.
This capability is derived from data without the need of hand-crafted rules.

In \cite{intentnet_corl18}, the authors propose a learnable end-to-end model
with a deep neural network that reasons about both high level behavior and
long-term trajectories. Inspired by how humans perform this task, the
network exploits motion and prior knowledge about the road topology in the
form of maps containing semantic elements such as lanes, intersections and
traffic lights. The so-called IntentNet is a fully-convolutional neural
network that outputs three types of variables in a single forward pass
corresponding to: detection scores for vehicle and background classes, high
level action probabilities corresponding to discrete intentions, and
bounding box regressions in the current and future time steps to represent
the intended trajectory. This design enables the system to propagate
uncertainty through the different components and is reported to be
computationally efficient.

A CNN is also used in \cite%
{Nikhil_Convolutional_Neural_Network_for_Trajectory_Prediction_ECCVW_2018_paper}
for an end-to-end trajectory prediction model which is competitive with more
complicated state-of-the-art LSTM-based techniques which require more
contextual information. Highly parallelizable convolutional layers are
employed to handle temporal dependencies. The CNN is a simple
sequence-to-sequence architecture. Trajectory histories are used as input
and embedded to a fixed size through a fully-connected layer. The
convolutional layers are stacked and used to enforce temporal consistency.
Finally, the features from the final convolutional layer are concatenated
and passed through a fully-connected layer to generate all predicted
positions at once. The authors found out that predicting one time step at a
time leads to worse results than predicting all future times at once. A
possible reason is that the error of the current prediction is propagated
forward in time in a highly correlated fashion.

\subsection{Methods Using Other Techniques}

The authors of \cite{18-zhou-ICRA} use Partially Observable Markov Decision
Processes (POMDPs) for behavior prediction and nonlinear receding horizon
control (or model predictive control) for trajectory planning. The POMDPs
model the interactions between the ego vehicle and the obstacles. The action
space is discretized into: acceleration, deceleration and maintaining the
current speed. For each of the obstacle vehicles, three types of intentions
are considered: going straight, turning and stopping. The reward function is
chosen so that the actors make the maximum progress on the road while
avoiding collisions. A particle filter is implemented to update the belief
of each motion intention for each obstacle vehicle. For the ego car, the
bicycle kinematic model is used to update the state.

Article \cite{2018_IV_0596} presents a simple yet effective way to
accurately predict the future trajectories of observed vehicles in dense
city environments. The authors recorded the trajectories of cars comprising
over 1000 hours of driving in San Francisco and New York. By relating the
current position of an observed car to this large dataset of previously
exhibited motion in the same area, the prediction of its future position can
be directly performed. Under the hypothesis that the car follows the same
trajectory pattern as one of the cars in the past at the same location had
followed. This non-parametric method improves over time as the amount of
samples increases and avoids the need for more complex models.

Paper \cite{IROS13_PIN_161867_} presents a trajectory prediction method
which combines the constant yaw rate and acceleration (CYRA) motion model
and maneuver recognition. The maneuver recognition module selects the
current maneuver from a predefined set (e.g. keep lane, change lane to the
right or to the left and turn at an intersection) by comparing the center
lines of the road lanes to a local curvilinear model of the path of the
vehicle. The proposed method combines the short-term accuracy of the former
technique and the longer-term accuracy of the latter. The authors use
mathematical models that take into account the position, speed and
acceleration of vehicles.

In \cite{root_pdf}, a method is presented that evaluates the probabilistic
prediction of real traffic scenes with varying start conditions. The
prediction is based on a particle filter, which estimates the
behavior-describing parameters of a microscopic traffic model, i.e. the
driving style as a distribution of behavior parameters. This method seems to
be applicable for long-term trajectory planning. The driving style
parameters of the intelligent driving model (IDM) are continuously
estimated, together with the relative motion between objects. By measuring
vehicle accelerations, a driving style estimation can be provided from the
first detection without the need of a long observation time before
performing the prediction. The use of a particle filter enables to cope with
continuous behavior changes with arbitrarily shaped parameter distributions.
Forward propagation using Monte Carlo simulation provides an approximate
probability density function of the future scene.

Since Markov models are only conditioned on the last observed position, they
can generate poor predictions if different motion patterns exhibit
significantly overlapping segments. Moreover, trajectories acquired from
sensors can be fragmented by occlusion. The approaches based on Gaussian
Processes (GPs) overcome this problem by modeling motion patterns as
velocity flow fields, thus avoiding the need to identify goal positions.
Also, they are well-suited for applications with noisy measurements, such as
data collected on moving cars. More importantly, predictions using a GP have
a simple analytical form that can be easily integrated into a risk-aware
path planner. Article \cite{scitech_gnc16} develops a data-driven approach
for learning a mobile agent’s motion patterns from past observations, which
are subsequently used for online trajectory predictions. It examines the
reasons why previous GP-based mixture models can sometimes produce poor
prediction results by providing examples to show that while GP is a flexible
tool for modeling motion patterns, GP likelihood is not a good similarity
measure for trajectory clustering.

As the traffic participants have a mutual influence on one another, their
interaction is explicitly considered in \cite{1216567}, which is inspired by
an optimization problem. For motion prediction, the collision probability of
a vehicle performing a certain maneuver is computed. The prediction is
performed based on the safety evaluation and the assumption that drivers
avoid collisions. This combination of the intention of each driver and the
driver's local risk assessment to perform a maneuver leads to an
interaction-aware motion prediction. The authors compute the probability
that a collision will occur anywhere in the whole scene, considering that
the number of different maneuvers is limited (e.g., lane changes,
acceleration, maintaining the speed, deceleration, and combinations), and
then the proposed system assesses the danger of possible future trajectories.

The same concept of considering risk is used in \cite%
{Integrated_Maneuver-Based_Trajectory}, which describes an integrated
Bayesian approach to maneuver-based trajectory prediction and criticality
assessment that is not limited to specific driving situations. First, a
distribution of high-level driving maneuvers is inferred for each vehicle in
the traffic scene by means of Bayesian inference. For this purpose, the
domain is modeled with a Bayesian network. Subsequently, maneuver-based
probabilistic trajectory prediction models are employed to predict the
configuration of each vehicle forward in time. The proposed system has three
main parts: the maneuver detection, the prediction, and the criticality
assessment. In the maneuver detection part, the current driving maneuver of
every vehicle is estimated via Bayesian inference. In the prediction part,
maneuver-specific prediction models are employed to predict the
configuration of each vehicle forward in time within a common global
coordinate system. In the criticality assessment part, these individual
joint distributions are used together with a parametric free space map-based
representation of the static environment with probability distribution
functions to estimate the collision probability of the event that the ego
vehicle collides with at least one other vehicle or the static driving
environment at least once within the prediction horizon via Monte Carlo
simulation.

The authors of \cite{1801.06523} propose a framework for holistic surround
vehicle trajectory prediction with three interacting modules: a trajectory
prediction module, based on the combination of an interaction model based on
motion and maneuver specific variational Gaussian mixture models, a maneuver
recognition module based on hidden Markov models for assigning confidence
values for maneuvers being performed by surrounding vehicles, and a vehicle
interaction module that considers the global context of surrounding vehicles
and assigns final predictions by minimizing an energy function based on
outputs of the other two modules. The motion model becomes unreliable for
long-term trajectory prediction, especially in cases involving a greater
degree of decision making by drivers. The paper defines ten maneuver classes
for surrounding vehicle motion on freeways in the frame of reference of the
ego vehicle, based on combinations of lane passes, overtakes, cut-ins and
drift into ego lane. A corresponding energy minimization problem is set so
that the predictions where at any point in the time horizon, two vehicles
are very close to each other, are penalized. This is based on the fact that
drivers tend to follow paths with low probability of collision with other
vehicles.

\subsection{Mixed Methods}

The authors of \cite{AAMAS_2019} take a model-based approach to prediction,
in order to make use of structured prior knowledge of vehicle kinematics,
and the assumption that other drivers plan trajectories to minimize an
unknown cost function. They introduce an IOC algorithm to learn the cost
functions of other vehicles in an energy-based generative model. Langevin
sampling, a Monte Carlo-based sampling algorithm, is used to directly sample
the control sequence. Langevin sampling is shown to generate better
predictions with higher stability. It seems that this algorithm is more
flexible that standard IOC methods, and can learn higher-level,
non-Markovian cost functions defined over entire trajectories. The weighted
feature-based cost functions are extended with neural networks in order to
combine the advantages of both model-based and model-free learning. The
study uses both environment structure (in the form of kinematic vehicular
constraints, which can be modeled very accurately), and the assumption that
human drivers optimize their trajectories according to a subjective cost
function. The results show that model-based IOC can achieve state-of-the-art
vehicle trajectory prediction accuracy and naturally take scene information
into account.

Multiple deep neural network architectures are designed to learn the cost
functions, some of which augment a set of hand-crafted features. The
human-crafted cost functions are defined as ten components: the distance to
the goal, the distance to the center of the lane, the penalty of collision
to other vehicles (inversely proportional to the distance to other
vehicles), the L2-norm of acceleration and steering, the L2-norm for the
difference of acceleration and steering between two frames, the heading
angle to lane, and the difference to the speed limit.

A thesis \cite{255338} investigates the application of deep learning and
mixture models for the prediction of human drivers in traffic. The chosen
approach is a mixture density network where the neural network is composed
of LSTM units and the mixture model consists of univariate Gaussian
distributions. It applies multi-task learning, in that by sharing the
representation between multiple tasks, one enables the model to generalize
better. A limitation is that the tasks usually have to be related to some
extent. For example, a single neural network can predict both longitudinal
and lateral accelerations from the same input, where the first few layers in
the network are shared between the two tasks, and then separated into two
different layers to produce the final outputs. To capture the intention of
the driver, another layer is used in parallel to the motion prediction layer
after the LSTM layers. This layer indicates if the driver intends to switch
lane and remain there within the next four seconds.

\subsection{Discussion}

Rule-based approaches to vehicle interaction are rather inflexible; they
require a great effort to engineer and validate, and they usually generalize
poorly to new scenarios \cite{AAMAS_2019}.

Learning-based approaches are promising because of the complexity of driving
interactions, and the need for generalization. However, learning-based
systems require a large amount of data to cover the space of interactive
behaviors. Because they capture the generative structure of vehicle
trajectories, model-based methods can potentially learn more, from less
data, than model-free methods. However, good cost functions are challenging
to learn, and simple, hand-crafted representations may not generalize well
across tasks and contexts. In general, model-based methods can be less
flexible, and may underperform model-free methods in the limit of infinite
data. Model-free methods take a data-driven approach, aiming to learn
predictive distributions over trajectories directly from data. These
approaches are more flexible and require less knowledge engineering in terms
of the type of vehicles, maneuvers, and scenarios, but the amount of data
they require may be prohibitive \cite{AAMAS_2019}.

Manually designed engineered models often impose unrealistic assumptions not
supported by the data, e.g., that traffic always follows lanes, which
motivated the use of learned models as an alternative. A large class of
learned models are maneuver-based models, e.g., using hidden Markov models,
which are object-centric approaches that predict the discrete actions of
each object independently. Often, the independence assumption is not true,
which is mitigated by the use of Bayesian networks that are computationally
more expensive and not feasible for real-time tasks \cite{1808.05819}.

Gaussian Process regression can also be used to address the motion
prediction problem. It has desirable properties such as the ability to
quantify uncertainty, but it is limited when modeling complex
actor-environment interactions \cite{1808.05819}.

Although it is possible to do multi-step prediction with a Kalman filter, it
cannot be extended far into the future with reasonable accuracy. A
multi-step prediction done solely by a Kalman filter was found to be
accurate up until 10-15 timesteps, after which the predictions diverged and
the full 40 timesteps prediction ended up being worse than constant velocity
inference \cite{255338}. This emphasizes the advantages of data-driven
approaches, as it is possible to observe almost an infinite number of
variables which may all affect the driver, whereas the Kalman filter relies
solely on the physical movement of the vehicle.

The data may also be a part of the problem, because the network learns what
is present in the data, and hopefully generalizes well, but there may always
be situations where the humans do not behave according to previous
observations. This is one drawback of using neural networks. However, it
seems that the advantages of using a data-driven approach outperform the
disadvatages.

Because of the time constraints of real-time systems, some authors use
simpler feed-forward CNN architectures for prediction \cite{1808.05819}. In
general, deep CNNs as robust, flexible, high-capacity function
approximators, are able to model the complex relationship between sensory
input and reward structure very well. Additionally, due to the convolutional
operators, they are able to capture spatial correlations in the data \cite%
{annurev-control-060117-105157}. Some authors \cite%
{Nikhil_Convolutional_Neural_Network_for_Trajectory_Prediction_ECCVW_2018_paper}
state that CNNs are superior to LSTMs for temporal modeling since
trajectories are continuous in nature, do not have complicated ``state'',
and have high spatial and temporal correlations which can be exploited by
computationally efficient convolution operations.

Another approach is to learn policies from expert demonstrations by
estimating the expert's cost function with inverse reinforcement learning
and then extract a policy from that cost function \cite%
{annurev-control-060117-105157}. However, this is often inefficient for
real-time applications \cite{1808.05819}.

Finally, it should be mentioned that in this section, we have addressed the
trajectory prediction problem. A related, but distinct problem, is
trajectory planning, i.e. finding an optimal path from the current location
to a given goal location. Its aim is to produce smooth trajectories with
small changes in curvature, so as to minimize both the lateral and the
longitudinal acceleration of the ego vehicle. For this purpose, there are
several methods reported in the literature, e.g. using cubic spline
interpolation, trigonometric spline interpolation, Bézier curves, or
clothoids, i.e. curves with a complex mathematical definition, which have a
linear relation between the curvature and the arc length and allow smooth
transitions from a straight line to a circle arc or vice versa.

\section{Decision Making Methods}

Since an agent's actions depend on the other agents' actions, an uncertainty
explosion in future states may arise and this may result in the
freezing-robot problem, where a robot comes to a complete stop because all
possible actions become unacceptably unsafe. If the robot does not come to a
complete stop, it may choose to follow highly evasive or arbitrary paths
through the problem space, which are often not only suboptimal but
potentially dangerous \cite{annurev-control-060117-105157}.

While modeling interactions is an intriguing problem in itself, dealing with
the increased complexity is another challenge. Since all agents' actions are
affected and equally affect other agents' actions, the number of
interactions (and therefore the planning complexity) grows exponentially
with the number of agents. The simplest approach is to discretize the action
space by motion primitives and to exhaustively search through all possible
options. Naturally, there are more efficient methods of exploring the
optimization space. In the deterministic case, one can cover the
decision-making process, often phrased in a game-theoretic setting, in a
tree-type structure and apply a search over the tree. The tree, usually
discretized by action time, consists of discrete actions that each agent can
choose to execute at each stage. Since each agent's reward depends not only
on its own reward and actions but also on all other agents' actions at the
previous stages, the tree also grows exponentially with the number of agents 
\cite{annurev-control-060117-105157}.

In the previous section, we have presented prediction methods for the
trajectories of surrounding vehicles. An important issue is related to the
decisions of the ego car itself regarding the possible maneuvers that it can
make in order to optimize some criteria related to risk and efficiency. In
this section, we briefly present some methods that can be used for this
purpose. We especially focus on (deep) reinforcement learning (RL) and tree
search algorithms.

\subsection{Deep Reinforcement Learning Algorithms}

Recently, there have been many efforts in devising better, more efficient RL
algorithms. A very popular class of applications is represented by games,
where the task is to learn to play directly from the game image and perhaps
the score, without any a priori knowledge about the game rules. Of course,
the same algorithms can be applied to other classes of problems, including
decision making in autonomous driving. Below, we present some of these RL
algorithms \cite{sutton}, \cite{lapan}:

\begin{itemize}
\item \textit{Policy Gradients} \cite{polgrad}. The objective of an RL agent
is to maximize the expected total discounted reward, i.e. value or utility,
by following a policy. The policy returns the action that the agent should
take in each state. This is usually a maximization problem (finding the best
action in every state) and the maximum function is not differentiable, so
gradient-based methods cannot be used. However, one can use a parametric
representation for the policy, e.g., a neural network that gives the
probabilities of each action for each state using the softmax function.
Softmax is differentiable, therefore gradients can be used to adjust the
parameters of the neural network which, in turn, approximates the policy;

\item \textit{Deep Q-Network (DQN)} \cite{dqn}. It approximates the \textit{Q%
} matrix of values computed, e.g., by the classic \textit{Q}-Learning
algorithm with a neural network. A great advantage is that each step of
experience is likely used in many weight updates, which allows better
generalization to unvisited states. However, it was found that learning
directly from successive samples is suboptimal because of the correlations
between the samples. Instead, the algorithm learns using experience replay,
i.e. the updates are made using random samples from a buffer of past
transitions. Also, in order to stabilize learning, the target network is
kept fixed for a certain number of learning episodes, and then replaced by
the current network;

\item \textit{Actor-Critic} \cite{actor_critic}. These methods are temporal
difference (TD) methods that have a separate memory structure to explicitly
represent the policy independently of the value function. The policy
structure is known as the ``actor'', because it is used to select actions,
and the estimated value function is known as the ``critic'', because it
criticizes the actions made by the actor. Learning is on-policy: the critic
learns about and critiques, in the form of a TD error, the policy followed
by the actor. This scalar signal is the only output of the critic and drives
all learning in both actor and critic;

\item \textit{Asynchronous Advantage Actor-Critic (A3C)} \cite{1602.01783}.
In A3C there is a global network and multiple worker agents each with its
own network. Each of these agents interacts with its own copy of the
environment at the same time. In this way, the experience of each agent is
independent of the experience of the others and thus the overall experience
available for training becomes more diverse. Instead of discounted rewards,
the method uses another value called ``advantage'', which allows the agent
to determine not just how good its actions were, but how much better they
turned out to be than expected. The advantage is positive if an action is
better than the other actions possible in that state;

\item \textit{Proximal Policy Optimization} \cite{ppo}. It improves the
stability of the actor training by limiting the policy update at each
training step. Thus, it avoids having too large policy updates. The ratio
that represents the difference between the new and the old policy is clipped
(e.g., between 0.8 and 1.2), ensuring that the policy updates are not too
large;

\item \textit{Trust Region Policy Optimization (TRPO)} \cite{trpo}. Policy
Gradients computes the steepest ascent direction for the rewards and updates
the policy towards that direction. However, this method uses the first-order
derivative and approximates the surface to be flat. If the surface has high
curvature and the step size (the learning rate) is too large, it can lead to
very bad policies. On the other hand, if the step is too small, the model
learns too slowly. TRPO limits the parameter changes that are sensitive to
the cost surface and ensures that any policy change should guarantee an
improvement in rewards. In the trust region, one determines the maximum step
size that is used for exploration and locates the optimal point within this
trust region. If the divergence between the new and the old policy is
getting large, the trust region is shrunk; otherwise, it is expanded;

\item \textit{Imagination-Augmented Agent} \cite{iaa}. The idea of this
algorithm is to allow the agent to imagine future trajectories and
incorporate these imagined paths into its decision process. They consist of
a set of trajectories ``imagined'' from the current observation. The
trajectories are called ``rollouts'' and are produced for every available
action in the environment. Every rollout consists of a fixed number of steps
into the future and every step in a special model called the ``environment
model'' produces the next observation and predicts the immediate reward from
the current observation and the action to be taken.
\end{itemize}

There are several papers that explore different variants of these algorithms.

The authors of \cite{1602.01783} propose a conceptually simple and
lightweight framework for deep reinforcement learning that uses asynchronous
gradient descent. They present asynchronous variants of four standard
reinforcement learning algorithms and show that parallel actor-learners have
a stabilizing effect on training, allowing all four methods to successfully
train neural network controllers. Instead of experience replay, multiple
agents are executed in parallel, on multiple instances of the environment.
This parallelism also decorrelates the agents' data into a more stationary
process. The experiments are run on a single machine with a standard
multi-core CPU. The best of the proposed methods is reported to be the A3C.

Article \cite{1611.06256} introduces a hybrid CPU/GPU version of the A3C
algorithm and concentrates on aspects critical to leveraging the
computational power of the GPU. It introduces a system of queues and a
dynamic scheduling strategy and achieves a significant speed up with respect
to its CPU equivalent.

The authors of \cite%
{CONTINUOUS_CONTROL_WITH_DEEP_REINFORCEMENT_LEARNING_-__1509.02971v5} adapt
the ideas underlying the success of Deep Q-Learning to the continuous action
domain. They present an actor-critic, model-free, off-policy (i.e. the
network is trained off-policy with samples from a replay buffer) algorithm
based on the deterministic policy gradient that can operate over continuous
action spaces. The actor-critic approach is combined with insights from DQN.
The resulting model seems to be able to learn competitive policies using
low-dimensional observations, e.g. Cartesian coordinates or joint angles. A
key feature of the approach is its simplicity: it requires only a
straightforward actor-critic architecture and learning algorithm with very
few adjustable parameters. Its main disadvantage is that it requires a large
number of training episodes to find solutions.

In \cite{jair1}, proximal gradient temporal difference learning is
introduced, which provides a principled way of designing and analyzing true
stochastic gradient temporal difference learning algorithms. The authors
show how gradient temporal difference (GTD) reinforcement learning methods
can be formally derived, not by starting from their original objective
functions, as previously attempted, but rather from a primal-dual
saddle-point objective function. Both error bound and performance bound are
provided, which shows that the value function approximation bound of the GTD
algorithms family is $O\left( {d/\sqrt[4]{n}} \right)$, where $d$ is the
dimension of the feature vector and $n$ is the number of samples.

Direct policy search can effectively scale to high-dimensional systems, but
complex policies with hundreds of parameters often present a challenge for
such methods, requiring numerous samples and often falling into poor local
optima. Article \cite{levine13} presents a guided policy search algorithm
that uses trajectory optimization to direct policy learning and avoid poor
local optima. It shows how differential dynamic programming can be used to
generate suitable guiding samples, and describes a regularized importance
sampled policy optimization that incorporates these guiding samples into the
policy search. As a consequence, the algorithm can learn complex policies
with hundreds of parameters.

Another interesting algorithm is the Predictron \cite{Predictron}. This
architecture is an abstract model, represented by a Markov reward process,
that can be rolled forward multiple ``imagined'' planning steps. Each
forward pass of the predictron accumulates internal rewards and values over
multiple planning depths. The predictron is trained end-to-end so as to make
these accumulated values accurately approximate the true value function. It
is reported to demonstrate more accurate predictions than conventional deep
neural network architectures. The predictron is composed of four main
components. First, a state representation that encodes raw input (e.g., a
history of observations, in the partially observed setting) into an internal
(abstract, hidden) state. Second, a model that maps from an internal state
to a subsequent internal state, internal reward, and internal discount.
Third, a value function that outputs internal values representing the
future, internal return from the internal state onwards. The predictron is
applied by unrolling its model multiple ``planning'' steps to produce
internal rewards, discounts and values. Finally, these internal rewards,
discounts and values are combined together by an accumulator into an overall
estimate of value. Unlike most approaches to model-based RL, the model is
fully abstract: it does not have to correspond to the real environment in
any human understandable fashion, as long as its rolled-forward ``plans''
accurately predict the outcomes in the true environment.

\subsection{Tree Search Algorithms}

Planning problems are often solved by tree search algorithms that simulate
ahead into the future, evaluate future states, and back-up those evaluations
to the root of the search tree. Among these algorithms, Monte Carlo Tree
Search (MCTS) \cite{mcts} is one of the most general, powerful and widely
used. The typical MCTS algorithm consists of several phases. First, it
simulates trajectories into the future, starting from the root state.
Second, it evaluates the performance of the leaf states, either using a
random rollout, or using an evaluation function such as a ``value network''.
Third, it backs-up these evaluations to update the internal values along the
trajectory, for example by averaging over evaluations.

The architecture presented in \cite{1802.04697}, called MCTSnet,
incorporates the simulation-based search into a neural network, by
expanding, evaluating and backing-up a vector embedding. The parameters of
the network are trained end-to-end using gradient-based optimization. The
key idea is to represent the internal state of the search, at each node, by
a memory vector. The computation of the network proceeds forwards from the
root state, just like a simulation of MCTS, using a simulation policy based
on the memory vector to select the trajectory to traverse. The leaf state is
then processed by an embedding network to initialize the memory vector at
the leaf. The network proceeds backwards up the trajectory, updating the
memory at each visited state according to a backup network that propagates
from child to parent. Finally, the root memory vector is used to compute an
overall prediction of value or action. The major benefit of this
architecture is that it can be used for gradient-based optimization. Still,
internal action sequences directing the control flow of the network cannot
be differentiated, and learning this internal policy presents a challenging
credit assignment problem. To address this, \cite{1802.04697} proposes a
novel, generally-applicable approximate scheme for credit assignment that
leverages the anytime property of the computational graph, allowing to
effectively learn this part of the search network from data.

Rapidly-exploring random trees (RRTs) \cite{rrt} represent an efficient
method for finding feasible trajectories for high-dimensional non-holonomic
systems. They can be viewed as a technique to generate open-loop
trajectories for nonlinear systems with state constraints. An RRT can also
be considered as a Monte Carlo method to bias search into the largest
Voronoi regions of a graph in a configuration space. The tree is constructed
incrementally from samples drawn randomly from the search space and is
inherently biased to grow towards large unsearched areas of the problem. If
the random sample is further from its nearest state in the tree than this
limit allows, a new state at the maximum distance from the tree along the
line to the random sample is used instead of the random sample itself. The
random samples can then be viewed as controlling the direction of the tree
growth while the growth factor determines its rate.

Article \cite{KuwataTCST09} describes a real-time motion planning algorithm,
based on RRTs, applicable to autonomous vehicles operating in an urban
environment. The extensions to the standard RRT are motivated by the need to
generate dynamically feasible plans in real-time, safety requirements, and
the constraints dictated by the uncertainty of driving in an urban
environment. The proposed algorithm was at the core of the planning and
control software for Team MIT's entry for the 2007 DARPA Urban Challenge,
where the vehicle demonstrated the ability to complete almost 100 km of a
simulated military supply mission, while safely interacting with other
autonomous and human driven vehicles.

\section{Conclusions}

The past three decades have seen increasingly rapid progress in driverless
vehicle technology. In addition to the advances in computing and perception
hardware, this rapid progress has been enabled by major theoretical progress
in computational aspects. Autonomous cars are complex systems which can be
decomposed into a hierarchy of decision making problems, where the solution
of one problem is the input to the next. The breakdown into individual
decision making problems has enabled the use of well-developed methods and
technologies from a variety of research areas \cite%
{2016ASurveyofMotionPlanningSelf-drivingUrbanVehicles}. This literature
review has concentrated only on three aspects: tracking, trajectory
prediction and decision making. It can serve as a reference for assessing
the computational tradeoffs between various choices for algorithm design.

\section*{Acknowledgements}

We kindly thank Continental AG for their great cooperation within \textit{Proreta 5}, 
which is a joint research project of the Technical University of Darmstadt, 
University of Bremen, Technical University of Ia\c{s}i and Continental AG.

\bibliographystyle{acm}
\bibliography{Tracking_Prediction_Refs1,Tracking_Prediction_Refs2}

\begin{thebibliography}{100}

\bibitem{7577010}
{\sc {Andersen}, H., {Chong}, Z.~J., {Eng}, Y.~H., {Pendleton}, S., and {Ang},
  M.~H.}
\newblock Geometric path tracking algorithm for autonomous driving in
  pedestrian environment.
\newblock In {\em 2016 IEEE International Conference on Advanced Intelligent
  Mechatronics (AIM)\/} (July 2016), pp.~1669--1674.

\bibitem{255338}
{\sc Andersson, J.}
\newblock Predicting vehicle motion and driver intent using deep learning.
\newblock Master's thesis, Chalmers University of Technology, G\"{o}teborg,
  Sweden, 2018.

\bibitem{safety-first-for-automated-driving}
{\sc Aptiv, Audi, Baidu, {BMW}, Continental, Daimler, Fiat, {Chrysler
  Automobiles}, {HERE}, Infineon, Intel, and Volkswagen}.
\newblock Safety first for automated driving, 2019.

\bibitem{1611.06256}
{\sc Babaeizadeh, M., Frosio, I., Tyree, S., Clemons, J., and Kautz, J.}
\newblock {GA3C:} gpu-based {A3C} for deep reinforcement learning.
\newblock {\em CoRR abs/1611.06256\/} (2016).

\bibitem{Bae_Robust_Online_Multi-Object_2014_CVPR_paper}
{\sc {Bae}, S., and {Yoon}, K.}
\newblock Robust online multi-object tracking based on tracklet confidence and
  online discriminative appearance learning.
\newblock In {\em 2014 IEEE Conference on Computer Vision and Pattern
  Recognition\/} (June 2014), pp.~1218--1225.

\bibitem{BerclazFTF11}
{\sc {Berclaz}, J., {Fleuret}, F., {Turetken}, E., and {Fua}, P.}
\newblock Multiple object tracking using k-shortest paths optimization.
\newblock {\em IEEE Transactions on Pattern Analysis and Machine Intelligence
  33}, 9 (Sep. 2011), 1806--1819.

\bibitem{1903.05625}
{\sc Bergmann, P., Meinhardt, T., and Leal-Taix{\'e}, L.}
\newblock Tracking without bells and whistles.
\newblock {\em ArXiv abs/1903.05625\/} (2019).

\bibitem{1602.00763}
{\sc {Bewley}, A., {Ge}, Z., {Ott}, L., {Ramos}, F., and {Upcroft}, B.}
\newblock Simple online and realtime tracking.
\newblock In {\em 2016 IEEE International Conference on Image Processing
  (ICIP)\/} (Sep. 2016), pp.~3464--3468.

\bibitem{Goutam_Bhat_Unveiling_the_Power_ECCV_2018_paper}
{\sc Bhat, G., Johnander, J., Danelljan, M., Khan, F.~S., and Felsberg, M.}
\newblock Unveiling the power of deep tracking.
\newblock In {\em ECCV\/} (2018).

\bibitem{1517Bochinski20176}
{\sc {Bochinski}, E., {Eiselein}, V., and {Sikora}, T.}
\newblock High-speed tracking-by-detection without using image information.
\newblock In {\em 2017 14th IEEE International Conference on Advanced Video and
  Signal Based Surveillance (AVSS)\/} (Aug 2017), pp.~1--6.

\bibitem{1704.07911}
{\sc Bojarski, M., Yeres, P., Choromanska, A., Choromanski, K., Firner, B.,
  Jackel, L.~D., and Muller, U.}
\newblock Explaining how a deep neural network trained with end-to-end learning
  steers a car.
\newblock {\em CoRR abs/1704.07911\/} (2017).

\bibitem{intentnet_corl18}
{\sc Casas, S., Luo, W., and Urtasun, R.}
\newblock {IntentNet}: Learning to predict intention from raw sensor data.
\newblock In {\em 2nd Annual Conference on Robot Learning, CoRL 2018,
  Z{\"{u}}rich, Switzerland, 29-31 October 2018, Proceedings\/} (2018),
  pp.~947--956.

\bibitem{1408.3304}
{\sc Chari, V., Lacoste-Julien, S., Laptev, I., and Sivic, J.}
\newblock On pairwise costs for network flow multi-object tracking.
\newblock {\em 2015 IEEE Conference on Computer Vision and Pattern Recognition
  (CVPR)\/} (2015), 5537--5545.

\bibitem{8296360}
{\sc {Chen}, L., {Ai}, H., {Shang}, C., {Zhuang}, Z., and {Bai}, B.}
\newblock Online multi-object tracking with convolutional neural networks.
\newblock In {\em 2017 IEEE International Conference on Image Processing
  (ICIP)\/} (Sep. 2017), pp.~645--649.

\bibitem{scitech_gnc16}
{\sc Chen, Y.~F., Liu, M., Liu, S.-Y., Miller, J., and How, J.}
\newblock Predictive modeling of pedestrian motion patterns with {Bayesian}
  nonparametrics.

\bibitem{1902.08231}
{\sc Chu, P., Fan, H., Tan, C.~C., and Ling, H.}
\newblock Online multi-object tracking with instance-aware tracker and dynamic
  model refreshment.
\newblock {\em 2019 IEEE Winter Conference on Applications of Computer Vision
  (WACV)\/} (2019), 161--170.

\bibitem{Chu_Online_Multi-Object_Tracking_ICCV_2017_paper}
{\sc Chu, Q., Ouyang, W., Li, H., Wang, X., Liu, B., and Yu, N.}
\newblock Online multi-object tracking using cnn-based single object tracker
  with spatial-temporal attention mechanism.
\newblock {\em 2017 IEEE International Conference on Computer Vision (ICCV)\/}
  (2017), 4846--4855.

\bibitem{1809.10732}
{\sc Cui, H., Radosavljevic, V., Chou, F., Lin, T., Nguyen, T., Huang, T.,
  Schneider, J., and Djuric, N.}
\newblock Multimodal trajectory predictions for autonomous driving using deep
  convolutional networks.
\newblock {\em CoRR abs/1809.10732\/} (2018).

\bibitem{cui18a}
{\sc Cui, Z., Lu, N., Jing, X., and Shi, X.}
\newblock Fast dynamic convolutional neural networks for visual tracking.
\newblock In {\em ACML\/} (2018).

\bibitem{ECO1611.09224}
{\sc Danelljan, M., Bhat, G., Khan, F.~S., and Felsberg, M.}
\newblock {ECO:} efficient convolution operators for tracking.
\newblock In {\em 2017 {IEEE} Conference on Computer Vision and Pattern
  Recognition, {CVPR} 2017, Honolulu, HI, USA, July 21-26, 2017\/} (2017),
  pp.~6931--6939.

\bibitem{ConvDCF_ICCV15_VOTworkshop}
{\sc {Danelljan}, M., {H\"{a}ger}, G., {Khan}, F.~S., and {Felsberg}, M.}
\newblock Convolutional features for correlation filter based visual tracking.
\newblock In {\em 2015 IEEE International Conference on Computer Vision
  Workshop (ICCVW)\/} (Dec 2015), pp.~621--629.

\bibitem{1608.03773CCOT}
{\sc Danelljan, M., Robinson, A., Khan, F.~S., and Felsberg, M.}
\newblock Beyond correlation filters: Learning continuous convolution operators
  for visual tracking.
\newblock In {\em Computer Vision - {ECCV} 2016 - 14th European Conference,
  Amsterdam, The Netherlands, October 11-14, 2016, Proceedings, Part {V}\/}
  (2016), pp.~472--488.

\bibitem{1801.06523}
{\sc Deo, N., Rangesh, A., and Trivedi, M.~M.}
\newblock How would surround vehicles move? a unified framework for maneuver
  classification and motion prediction.
\newblock {\em IEEE Transactions on Intelligent Vehicles 3\/} (2018), 129--140.

\bibitem{1805.05499}
{\sc Deo, N., and Trivedi, M.~M.}
\newblock Multi-modal trajectory prediction of surrounding vehicles with
  maneuver based lstms.
\newblock {\em 2018 IEEE Intelligent Vehicles Symposium (IV)\/} (2018),
  1179--1184.

\bibitem{2017_IJRR_Dequaire}
{\sc Dequaire, J., Ondruska, P., Rao, D., Wang, D.~Z., and Posner, I.}
\newblock Deep tracking in the wild: End-to-end tracking using recurrent neural
  networks.
\newblock {\em I. J. Robotics Res. 37\/} (2018), 492--512.

\bibitem{Dicle_The_Way_They_2013_ICCV_paper}
{\sc {Dicle}, C., {Camps}, O.~I., and {Sznaier}, M.}
\newblock The way they move: Tracking multiple targets with similar appearance.
\newblock In {\em 2013 IEEE International Conference on Computer Vision\/} (Dec
  2013), pp.~2304--2311.

\bibitem{1808.05819}
{\sc Djuric, N., Radosavljevic, V., Cui, H., Nguyen, T., Chou, F., Lin, T., and
  Schneider, J.}
\newblock Motion prediction of traffic actors for autonomous driving using deep
  convolutional networks.
\newblock {\em CoRR abs/1808.05819\/} (2018).

\bibitem{sensors-19-00387}
{\sc Du, M., Ding, Y., Meng, X., Wei, H.-L., and Zhao, Y.}
\newblock Distractor-aware deep regression for visual tracking.
\newblock In {\em Sensors\/} (2019).

\bibitem{Learning_to_Detect_and_Track_Visible_and_Occluded}
{\sc Fabbri, M., Lanzi, F., Calderara, S., Palazzi, A., Vezzani, R., and
  Cucchiara, R.}
\newblock Learning to detect and track visible and occluded body joints in a
  virtual world.
\newblock In {\em Computer Vision - {ECCV} 2018 - 15th European Conference,
  Munich, Germany, September 8-14, 2018, Proceedings, Part {IV}\/} (2018),
  pp.~450--466.

\bibitem{SANet}
{\sc Fan, H., and Ling, H.}
\newblock Sanet: Structure-aware network for visual tracking.
\newblock {\em 2017 IEEE Conference on Computer Vision and Pattern Recognition
  Workshops (CVPRW)\/} (2016), 2217--2224.

\bibitem{Feichtenhofer_Detect_to_Track_ICCV_2017_paper}
{\sc Feichtenhofer, C., Pinz, A., and Zisserman, A.}
\newblock Detect to track and track to detect.
\newblock {\em 2017 IEEE International Conference on Computer Vision (ICCV)\/}
  (2017), 3057--3065.

\bibitem{1901.06129}
{\sc Feng, W., Hu, Z., Wu, W., Yan, J., and Ouyang, W.}
\newblock Multi-object tracking with multiple cues and switcher-aware
  classification.
\newblock {\em CoRR abs/1901.06129\/} (2019).

\bibitem{1803.03347}
{\sc Fernando, T., Denman, S., Sridharan, S., and Fookes, C.}
\newblock Tracking by prediction: A deep generative model for multi-person
  localisation and tracking.
\newblock In {\em IEEE Winter Conference on Applications of Computer Vision
  (WACV 2018)\/} (Lake Tahoe, NV, 2018), IEEE, pp.~1122--1132.

\bibitem{polgrad}
{\sc {Grondman}, I., {Busoniu}, L., {Lopes}, G. A.~D., and {Babuska}, R.}
\newblock A survey of actor-critic reinforcement learning: Standard and natural
  policy gradients.
\newblock {\em IEEE Transactions on Systems, Man, and Cybernetics, Part C
  (Applications and Reviews) 42}, 6 (Nov 2012), 1291--1307.

\bibitem{1802.04697}
{\sc Guez, A., Weber, T., Antonoglou, I., Simonyan, K., Vinyals, O., Wierstra,
  D., Munos, R., and Silver, D.}
\newblock Learning to search with {MCTSnets}.
\newblock {\em CoRR abs/1802.04697\/} (2018).

\bibitem{DeepTracking}
{\sc Hahn, M., Chen, S., and Dehghan, A.}
\newblock Deep tracking: Visual tracking using deep convolutional networks.
\newblock {\em CoRR abs/1512.03993\/} (2015).

\bibitem{root_pdf}
{\sc {Hoermann}, S., {Stumper}, D., and {Dietmayer}, K.}
\newblock Probabilistic long-term prediction for autonomous vehicles.
\newblock In {\em 2017 IEEE Intelligent Vehicles Symposium (IV)\/} (June 2017),
  pp.~237--243.

\bibitem{IROS13_PIN_161867_}
{\sc {Houenou}, A., {Bonnifait}, P., {Cherfaoui}, V., and {Yao}, W.}
\newblock Vehicle trajectory prediction based on motion model and maneuver
  recognition.
\newblock In {\em 2013 IEEE/RSJ International Conference on Intelligent Robots
  and Systems\/} (Nov 2013), pp.~4363--4369.

\bibitem{279}
{\sc Jiang, X., Zhen, X., Zhang, B., Yang, J., and Cao, X.}
\newblock Deep collaborative tracking networks.
\newblock In {\em BMVC\/} (2018).

\bibitem{67061}
{\sc Kampker, A., Sefati, M., Rachman, A. S.~A., Kreisk{\"o}ther, K., and
  Campoy, P.}
\newblock Towards multi-object detection and tracking in urban scenario under
  uncertainties.
\newblock {\em ArXiv abs/1801.02686\/} (2018).

\bibitem{1704.07049}
{\sc Kim, B., Kang, C.~M., Kim, J., Lee, S.-H., Chung, C.~C., and Choi, J.~W.}
\newblock Probabilistic vehicle trajectory prediction over occupancy grid map
  via recurrent neural network.
\newblock 399--404.

\bibitem{MHTR_ICCV2015}
{\sc {Kim}, C., {Li}, F., {Ciptadi}, A., and {Rehg}, J.~M.}
\newblock Multiple hypothesis tracking revisited.
\newblock In {\em 2015 IEEE International Conference on Computer Vision
  (ICCV)\/} (Dec 2015), pp.~4696--4704.

\bibitem{Chanho_Kim_Multi-object_Tracking_with_ECCV_2018_paper}
{\sc Kim, C., Li, F., and Rehg, J.~M.}
\newblock Multi-object tracking with neural gating using bilinear lstm.
\newblock In {\em ECCV\/} (2018).

\bibitem{actor_critic}
{\sc Konda, V.~R., and Tsitsiklis, J.~N.}
\newblock Actor-critic algorithms.
\newblock 2000, pp.~1008--1014.

\bibitem{KuwataTCST09}
{\sc {Kuwata}, Y., {Teo}, J., {Fiore}, G., {Karaman}, S., {Frazzoli}, E., and
  {How}, J.~P.}
\newblock Real-time motion planning with applications to autonomous urban
  driving.
\newblock {\em IEEE Transactions on Control Systems Technology 17}, 5 (Sep.
  2009), 1105--1118.

\bibitem{lapan}
{\sc Lapan, M.}
\newblock {\em Deep Reinforcement Learning Hands-On}.
\newblock Packt Publishing, 2018.

\bibitem{rrt}
{\sc Lavalle, S.~M.}
\newblock Rapidly-exploring random trees: A new tool for path planning.
\newblock Tech. rep., Computer Science Department, Iowa State University, 1998.

\bibitem{1216567}
{\sc {Lawitzky}, A., {Althoff}, D., {Passenberg}, C.~F., {Tanzmeister}, G.,
  {Wollherr}, D., and {Buss}, M.}
\newblock Interactive scene prediction for automotive applications.
\newblock In {\em 2013 IEEE Intelligent Vehicles Symposium (IV)\/} (June 2013),
  pp.~1028--1033.

\bibitem{DESIRE}
{\sc Lee, N., Choi, W., Vernaza, P., Choy, C.~B., Torr, P. H.~S., and
  Chandraker, M.~K.}
\newblock {DESIRE:} distant future prediction in dynamic scenes with
  interacting agents.
\newblock {\em CoRR abs/1704.04394\/} (2017).

\bibitem{s40648-014-0001-z}
{\sc Lef\`{e}vre, S., Vasquez, D., and Laugier, C.}
\newblock A survey on motion prediction and risk assessment for intelligent
  vehicles.
\newblock {\em Robomech Journal\/} (2014).

\bibitem{levine13}
{\sc Levine, S., and Koltun, V.}
\newblock Guided policy search.
\newblock In {\em 30th International Conference on Machine Learning, ICML
  2013\/} (06 2013).

\bibitem{301}
{\sc Li, K., Kong, Y., and Fu, Y.}
\newblock Multi-stream deep similarity learning networks for visual tracking.
\newblock In {\em Proceedings of the 26th International Joint Conference on
  Artificial Intelligence\/} (2017), IJCAI'17, AAAI Press, pp.~2166--2172.

\bibitem{2018111616343061}
{\sc Li, S., Ma, B., Chang, H., Shan, S., and Chen, X.}
\newblock Continuity-discrimination convolutional neural network for visual
  object tracking.
\newblock {\em 2018 IEEE International Conference on Multimedia and Expo
  (ICME)\/} (2018), 1--6.

\bibitem{CONTINUOUS_CONTROL_WITH_DEEP_REINFORCEMENT_LEARNING_-__1509.02971v5}
{\sc Lillicrap, T.~P., Hunt, J.~J., Pritzel, A., Heess, N., Erez, T., Tassa,
  Y., Silver, D., and Wierstra, D.}
\newblock Continuous control with deep reinforcement learning, 2015.

\bibitem{jair1}
{\sc Liu, B., Ghavamzadeh, M., Gemp, I., Liu, J., Mahadevan, S., and Petrik,
  M.}
\newblock Proximal gradient temporal difference learning: Stable reinforcement
  learning with polynomial sample complexity.
\newblock {\em Journal of Artificial Intelligence Research 63\/} (11 2018),
  461--494.

\bibitem{16Liu}
{\sc Liu, Q., Lu, X., He, Z., Zhang, C., and Chen, W.-S.}
\newblock Deep convolutional neural networks for thermal infrared object
  tracking.
\newblock {\em Knowledge-Based Systems\/} (07 2017).

\bibitem{1804.04555}
{\sc {Ma}, C., {Yang}, C., {Yang}, F., {Zhuang}, Y., {Zhang}, Z., {Jia}, H.,
  and {Xie}, X.}
\newblock Trajectory factory: Tracklet cleaving and re-connection by deep
  siamese bi-gru for multiple object tracking.
\newblock In {\em 2018 IEEE International Conference on Multimedia and Expo
  (ICME)\/} (July 2018), pp.~1--6.

\bibitem{Maksai_Non-Markovian_Globally_Consistent_ICCV_2017_paper}
{\sc {Maksai}, A., {Wang}, X., {Fleuret}, F., and {Fua}, P.}
\newblock Non-markovian globally consistent multi-object tracking.
\newblock In {\em 2017 IEEE International Conference on Computer Vision
  (ICCV)\/} (Oct 2017), pp.~2563--2573.

\bibitem{8443497}
{\sc {Manjunath}, A., {Liu}, Y., {Henriques}, B., and {Engstle}, A.}
\newblock Radar based object detection and tracking for autonomous driving.
\newblock In {\em 2018 IEEE MTT-S International Conference on Microwaves for
  Intelligent Mobility (ICMIM)\/} (April 2018), pp.~1--4.

\bibitem{pedestrianTrajectories}
{\sc {Møgelmose}, A., {Trivedi}, M.~M., and {Moeslund}, T.~B.}
\newblock Trajectory analysis and prediction for improved pedestrian safety:
  Integrated framework and evaluations.
\newblock In {\em 2015 IEEE Intelligent Vehicles Symposium (IV)\/} (June 2015),
  pp.~330--335.

\bibitem{aaai2017-anton-rnntracking}
{\sc Milan, A., Rezatofighi, S.~H., Dick, A., Reid, I., and Schindler, K.}
\newblock Online multi-target tracking using recurrent neural networks.
\newblock In {\em AAAI\/} (February 2017).

\bibitem{pami2014-anton}
{\sc Milan, A., Roth, S., and Schindler, K.}
\newblock Continuous energy minimization for multitarget tracking.
\newblock {\em IEEE TPAMI 36}, 1 (2014), 58--72.

\bibitem{1602.01783}
{\sc Mnih, V., Badia, A.~P., Mirza, M., Graves, A., Lillicrap, T.~P., Harley,
  T., Silver, D., and Kavukcuoglu, K.}
\newblock Asynchronous methods for deep reinforcement learning, 2016.

\bibitem{dqn}
{\sc Mnih, V., Kavukcuoglu, K., Silver, D., Rusu, A.~A., Veness, J., Bellemare,
  M.~G., Graves, A., Riedmiller, M., Fidjeland, A.~K., Ostrovski, G., Petersen,
  S., Beattie, C., Sadik, A., Antonoglou, I., King, H., Kumaran, D., Wierstra,
  D., Legg, S., and Hassabis, D.}
\newblock Human-level control through deep reinforcement learning.
\newblock {\em Nature 518}, 7540 (2015), 529--533.

\bibitem{ICIPReza2018_v3}
{\sc {Mozhdehi}, R.~J., {Reznichenko}, Y., {Siddique}, A., and {Medeiros}, H.}
\newblock Deep convolutional particle filter with adaptive correlation maps for
  visual tracking.
\newblock In {\em 2018 25th IEEE International Conference on Image Processing
  (ICIP)\/} (Oct 2018), pp.~798--802.

\bibitem{CVIU2016}
{\sc Naiel, M.~A., Ahmad, M.~O., Swamy, M., Lim, J., and Yang, M.-H.}
\newblock Online multi-object tracking via robust collaborative model and
  sample selection.
\newblock {\em Computer Vision and Image Understanding 154\/} (2017), 94 --
  107.

\bibitem{1510.07945MDnet}
{\sc Nam, H., and Han, B.}
\newblock Learning multi-domain convolutional neural networks for visual
  tracking.
\newblock {\em 2016 IEEE Conference on Computer Vision and Pattern Recognition
  (CVPR)\/} (2015), 4293--4302.

\bibitem{Nikhil_Convolutional_Neural_Network_for_Trajectory_Prediction_ECCVW_2018_paper}
{\sc Nikhil, N., and Morris, B.~T.}
\newblock Convolutional neural network for trajectory prediction.
\newblock {\em CoRR abs/1809.00696\/} (2018).

\bibitem{Seeing_beyond_seeing}
{\sc Ondr\'{u}\v{s}ka, P., and Posner, I.}
\newblock Deep tracking: Seeing beyond seeing using recurrent neural networks.
\newblock In {\em Proceedings of the Thirtieth AAAI Conference on Artificial
  Intelligence\/} (2016), AAAI'16, AAAI Press, pp.~3361--3367.

\bibitem{2016ASurveyofMotionPlanningSelf-drivingUrbanVehicles}
{\sc Paden, B., C{\'{a}}p, M., Yong, S.~Z., Yershov, D.~S., and Frazzoli, E.}
\newblock A survey of motion planning and control techniques for self-driving
  urban vehicles.
\newblock {\em CoRR abs/1604.07446\/} (2016).

\bibitem{greedy_fahim_albert}
{\sc {Pirsiavash}, H., {Ramanan}, D., and {Fowlkes}, C.~C.}
\newblock Globally-optimal greedy algorithms for tracking a variable number of
  objects.
\newblock In {\em CVPR 2011\/} (June 2011), pp.~1201--1208.

\bibitem{7463-deep-attentive-tracking-via-reciprocative-learning}
{\sc Pu, S., Song, Y., Ma, C., Zhang, H., and Yang, M.-H.}
\newblock Deep attentive tracking via reciprocative learning.
\newblock In {\em Proceedings of the 32Nd International Conference on Neural
  Information Processing Systems\/} (USA, 2018), NIPS'18, Curran Associates
  Inc., pp.~1935--1945.

\bibitem{cvpr16_hedge_tracking}
{\sc {Qi}, Y., {Zhang}, S., {Qin}, L., {Yao}, H., {Huang}, Q., {Lim}, J., and
  {Yang}, M.}
\newblock Hedged deep tracking.
\newblock In {\em 2016 IEEE Conference on Computer Vision and Pattern
  Recognition (CVPR)\/} (June 2016), pp.~4303--4311.

\bibitem{1802.08755}
{\sc Rangesh, A., and Trivedi, M.~M.}
\newblock No blind spots: Full-surround multi-object tracking for autonomous
  vehicles using cameras and lidars.
\newblock {\em ArXiv abs/1802.08755\/} (2019).

\bibitem{Liangliang_Ren_Collaborative_Deep_Reinforcement_ECCV_2018_paper}
{\sc Ren, L., Lu, J., Wang, Z., Tian, Q., and Zhou, J.}
\newblock Collaborative deep reinforcement learning for multi-object tracking.
\newblock In {\em The European Conference on Computer Vision (ECCV)\/}
  (September 2018).

\bibitem{ristani2016MTMC}
{\sc Ristani, E., Solera, F., Zou, R.~S., Cucchiara, R., and Tomasi, C.}
\newblock Performance measures and a data set for multi-target, multi-camera
  tracking.
\newblock {\em ArXiv abs/1609.01775\/} (2016).

\bibitem{ristaniAccv14}
{\sc Ristani, E., and Tomasi, C.}
\newblock Tracking multiple people online and in real time.
\newblock In {\em ACCV\/} (2014).

\bibitem{1803.10859}
{\sc {Ristani}, E., and {Tomasi}, C.}
\newblock Features for multi-target multi-camera tracking and
  re-identification.
\newblock In {\em 2018 IEEE/CVF Conference on Computer Vision and Pattern
  Recognition\/} (June 2018), pp.~6036--6046.

\bibitem{GMCP-Tracker_ECCV12}
{\sc Roshan~Zamir, A., Dehghan, A., and Shah, M.}
\newblock Gmcp-tracker: Global multi-object tracking using generalized minimum
  clique graphs.
\newblock In {\em Computer Vision -- ECCV 2012\/} (Berlin, Heidelberg, 2012),
  A.~Fitzgibbon, S.~Lazebnik, P.~Perona, Y.~Sato, and C.~Schmid, Eds., Springer
  Berlin Heidelberg, pp.~343--356.

\bibitem{1701.01909}
{\sc Sadeghian, A., Alahi, A., and Savarese, S.}
\newblock Tracking the untrackable: Learning to track multiple cues with
  long-term dependencies.
\newblock {\em 2017 IEEE International Conference on Computer Vision (ICCV)\/}
  (2017), 300--311.

\bibitem{Integrated_Maneuver-Based_Trajectory}
{\sc {Schreier}, M., {Willert}, V., and {Adamy}, J.}
\newblock An integrated approach to maneuver-based trajectory prediction and
  criticality assessment in arbitrary road environments.
\newblock {\em IEEE Transactions on Intelligent Transportation Systems 17}, 10
  (Oct 2016), 2751--2766.

\bibitem{trpo}
{\sc Schulman, J., Levine, S., Moritz, P., Jordan, M.~I., and Abbeel, P.}
\newblock Trust region policy optimization.
\newblock {\em CoRR abs/1502.05477\/} (2015).

\bibitem{ppo}
{\sc Schulman, J., Wolski, F., Dhariwal, P., Radford, A., and Klimov, O.}
\newblock Proximal policy optimization algorithms.
\newblock {\em CoRR abs/1707.06347\/} (2017).

\bibitem{Schulter_Deep_Network_Flow_CVPR_2017_paper}
{\sc Schulter, S., Vernaza, P., Choi, W., and Chandraker, M.~K.}
\newblock Deep network flow for multi-object tracking.
\newblock {\em 2017 IEEE Conference on Computer Vision and Pattern Recognition
  (CVPR)\/} (2017), 2730--2739.

\bibitem{annurev-control-060117-105157}
{\sc Schwarting, W., Alonso-Mora, J., and Rus, D.}
\newblock Planning and decision-making for autonomous vehicles.
\newblock vol.~1, pp.~187--210.

\bibitem{1802.09298}
{\sc Sharma, S., Ansari, J.~A., Murthy, J.~K., and Krishna, K.~M.}
\newblock Beyond pixels: Leveraging geometry and shape cues for online
  multi-object tracking.
\newblock {\em 2018 IEEE International Conference on Robotics and Automation
  (ICRA)\/} (2018), 3508--3515.

\bibitem{mcts}
{\sc Silver, D., Huang, A., Maddison, C.~J., Guez, A., Sifre, L., van~den
  Driessche, G., Schrittwieser, J., Antonoglou, I., Panneershelvam, V.,
  Lanctot, M., Dieleman, S., Grewe, D., Nham, J., Kalchbrenner, N., Sutskever,
  I., Lillicrap, T., Leach, M., Kavukcuoglu, K., Graepel, T., and Hassabis, D.}
\newblock Mastering the game of {Go} with deep neural networks and tree search.
\newblock {\em Nature 529}, 7587 (Jan. 2016), 484--489.

\bibitem{Predictron}
{\sc Silver, D., van Hasselt, H., Hessel, M., Schaul, T., Guez, A., Harley, T.,
  Dulac{-}Arnold, G., Reichert, D.~P., Rabinowitz, N.~C., Barreto, A., and
  Degris, T.}
\newblock The {Predictron}: End-to-end learning and planning.
\newblock {\em CoRR abs/1612.08810\/} (2016).

\bibitem{Prediction_in_Autonomous_Vehicle_-_TDS}
{\sc Singh, A.}
\newblock Prediction in autonomous vehicle - all you need to know, 2018.

\bibitem{7410854}
{\sc Solera, F., Calderara, S., and Cucchiara, R.}
\newblock Learning to divide and conquer for online multi-target tracking.
\newblock In {\em Proceedings of the 2015 IEEE International Conference on
  Computer Vision (ICCV)\/} (Washington, DC, USA, 2015), ICCV '15, IEEE
  Computer Society, pp.~4373--4381.

\bibitem{Son_Multi-Object_Tracking_With_CVPR_2017_paper}
{\sc {Son}, J., {Baek}, M., {Cho}, M., and {Han}, B.}
\newblock Multi-object tracking with quadruplet convolutional neural networks.
\newblock In {\em 2017 IEEE Conference on Computer Vision and Pattern
  Recognition (CVPR)\/} (July 2017), pp.~3786--3795.

\bibitem{iccv17b}
{\sc Song, Y., Ma, C., Gong, L., Zhang, J., Lau, R. W.~H., and Yang, M.-H.}
\newblock Crest: Convolutional residual learning for visual tracking.
\newblock In {\em 2017 IEEE International Conference on Computer Vision
  (ICCV)\/} (2017), pp.~2574--2583.

\bibitem{1810.1178}
{\sc Sun, S., Akhtar, N., Song, H., Mian, A.~S., and Shah, M.}
\newblock Deep affinity network for multiple object tracking.
\newblock {\em CoRR abs/1810.11780\/} (2018).

\bibitem{2018_IV_0596}
{\sc {Suraj}, M.~S., {Grimmett}, H., {Platinsk\'{y}}, L., and
  {Ondr\'{u}\v{s}ka}, P.}
\newblock Predicting trajectories of vehicles using large-scale motion priors.
\newblock In {\em 2018 IEEE Intelligent Vehicles Symposium (IV)\/} (June 2018),
  pp.~1639--1644.

\bibitem{sutton}
{\sc Sutton, R.~S., and Barto, A.~G.}
\newblock {\em Reinforcement Learning}.
\newblock MIT Press, 2018.

\bibitem{ICCV17RobustObjectTrackingBasedOnTemporalAndSpatialDeepNetworks}
{\sc {Teng}, Z., {Xing}, J., {Wang}, Q., {Lang}, C., {Feng}, S., and {Jin}, Y.}
\newblock Robust object tracking based on temporal and spatial deep networks.
\newblock In {\em 2017 IEEE International Conference on Computer Vision
  (ICCV)\/} (Oct 2017), pp.~1153--1162.

\bibitem{deeplk-icra}
{\sc Wang, C., Galoogahi, H.~K., Lin, C.-H., and Lucey, S.}
\newblock {Deep-LK} for efficient adaptive object tracking.
\newblock {\em 2018 IEEE International Conference on Robotics and Automation
  (ICRA)\/} (2018), 627--634.

\bibitem{FULLTEXT01_mp}
{\sc Ward, E.}
\newblock {\em Models Supporting Trajectory Planning in Autonomous Vehicles}.
\newblock PhD thesis, KTH Royal Institute of Technology, Stockholm, Sweden,
  2018.

\bibitem{iaa}
{\sc Weber, T., Racani{\`{e}}re, S., Reichert, D.~P., Buesing, L., Guez, A.,
  Rezende, D.~J., Badia, A.~P., Vinyals, O., Heess, N., Li, Y., Pascanu, R.,
  Battaglia, P., Silver, D., and Wierstra, D.}
\newblock Imagination-augmented agents for deep reinforcement learning.
\newblock {\em CoRR abs/1707.06203\/} (2017).

\bibitem{aaai19a}
{\sc Wen, L., Du, D., Li, S., Bian, X., and Lyu, S.}
\newblock Learning non-uniform hypergraph for multi-object tracking.
\newblock {\em CoRR abs/1812.03621\/} (2018).

\bibitem{CVPR2014_HyperGraphMultiTargetsTracker}
{\sc {Wen}, L., {Li}, W., {Yan}, J., {Lei}, Z., {Yi}, D., and {Li}, S.~Z.}
\newblock Multiple target tracking based on undirected hierarchical relation
  hypergraph.
\newblock In {\em 2014 IEEE Conference on Computer Vision and Pattern
  Recognition\/} (June 2014), pp.~1282--1289.

\bibitem{1703.07402}
{\sc {Wojke}, N., {Bewley}, A., and {Paulus}, D.}
\newblock Simple online and realtime tracking with a deep association metric.
\newblock In {\em 2017 IEEE International Conference on Image Processing
  (ICIP)\/} (Sep. 2017), pp.~3645--3649.

\bibitem{E371Final}
{\sc Woo, H., Sugimoto, M., Wu, J., Tamura, Y., Yamashita, A., and Asama, H.}
\newblock Trajectory prediction of surrounding vehicles using {LSTM} network.
\newblock In {\em 2013 IEEE Intelligent Vehicles Symposium (IV)\/} (2018).

\bibitem{7410891}
{\sc {Xiang}, Y., {Alahi}, A., and {Savarese}, S.}
\newblock Learning to track: Online multi-object tracking by decision making.
\newblock In {\em 2015 IEEE International Conference on Computer Vision
  (ICCV)\/} (Dec 2015), pp.~4705--4713.

\bibitem{AAMAS_2019}
{\sc Xu, Y., Zhao, T., Baker, C., Zhao, Y., and Wu, Y.~N.}
\newblock Learning trajectory prediction with continuous inverse optimal
  control via {Langevin} sampling of energy-based models.
\newblock {\em CoRR abs/1904.05453\/} (2019).

\bibitem{DLST_MM}
{\sc Yang, L., Liu, R., Zhang, D., and Zhang, L.}
\newblock Deep location-specific tracking.
\newblock In {\em Proceedings of the 25th ACM International Conference on
  Multimedia\/} (New York, NY, USA, 2017), MM '17, ACM, pp.~1309--1317.

\bibitem{1805.10916}
{\sc Yoon, Y.-C., Boragule, A., Yoon, K., and Jeon, M.}
\newblock Online multi-object tracking with historical appearance matching and
  scene adaptive detection filtering.
\newblock {\em 2018 15th IEEE International Conference on Advanced Video and
  Signal Based Surveillance (AVSS)\/} (2018), 1--6.

\bibitem{18-zhou-ICRA}
{\sc {Zhou}, B., {Schwarting}, W., {Rus}, D., and {Alonso-Mora}, J.}
\newblock Joint multi-policy behavior estimation and receding-horizon
  trajectory planning for automated urban driving.
\newblock In {\em 2018 IEEE International Conference on Robotics and Automation
  (ICRA)\/} (May 2018), pp.~2388--2394.

\bibitem{paper-deeptam}
{\sc Zhou, H., Ummenhofer, B., and Brox, T.}
\newblock Deeptam: Deep tracking and mapping.
\newblock In {\em Computer Vision -- ECCV 2018\/} (Cham, 2018), V.~Ferrari,
  M.~Hebert, C.~Sminchisescu, and Y.~Weiss, Eds., Springer International
  Publishing, pp.~851--868.

\bibitem{eccv2018_mot}
{\sc Zhu, J., Yang, H., Liu, N., Kim, M., Zhang, W., and Yang, M.-H.}
\newblock Online multi-object tracking with dual matching attention networks.
\newblock In {\em Computer Vision -- ECCV 2018\/} (Cham, 2018), Springer
  International Publishing, pp.~379--396.

\end{thebibliography}

\end{document}